\documentclass{article}

\usepackage{microtype}
\usepackage{graphicx}
\usepackage{subfigure}
\usepackage{booktabs} 
\usepackage{amsmath}
\usepackage{xcolor}
\usepackage{algpseudocode} 
\usepackage{enumitem}
\usepackage{flushend}
\usepackage{xspace}
\newcommand{\projectname}{LC-Checkpoint\xspace}

\providecommand{\mypara}[1]{\smallskip\noindent\emph{#1} }
\providecommand{\myparab}[1]{\smallskip\noindent\textbf{#1} }

\newcommand{\E}{\mathrm{I\! E}}

\newcommand{\mmu}{\mathbf{u}}

\newcommand{\mmw}{\mathbf{w}}

\newcommand{\reals}{\mathbf{R}}
\providecommand{\mypara}[1]{\smallskip\noindent\emph{#1} }
\providecommand{\myparab}[1]{\smallskip\noindent\textbf{#1} }

\usepackage{hyperref}

\usepackage[accepted]{icml2020}

\icmltitlerunning{On Efficient Constructions of Checkpoints}

\begin{document}

\twocolumn[
\icmltitle{On Efficient Constructions of Checkpoints}



\icmlsetsymbol{equal}{*}

\begin{icmlauthorlist}
\icmlauthor{Yu Chen}{to}
\icmlauthor{Zhenming Liu}{to}
\icmlauthor{Bin Ren}{to}
\icmlauthor{Xin Jin}{goo}
\end{icmlauthorlist}

\icmlaffiliation{to}{William \& Mary, Williamsburg, Virginia, USA}
\icmlaffiliation{goo}{Johns Hopkins University, Baltimore, Maryland, USA}

\icmlcorrespondingauthor{Yu Chen}{ychen39@email.wm.edu}


\icmlkeywords{checkpoint, compression}

\vskip 0.3in
]



\printAffiliationsAndNotice{}  

\begin{abstract}
Efficient construction of checkpoints/snapshots is a critical tool for training and diagnosing deep learning models. In this paper, we propose a lossy compression scheme for checkpoint constructions (called \projectname). \projectname simultaneously maximizes the compression rate and optimizes the recovery speed, under the assumption that SGD is used to train the model. \projectname uses quantization and priority promotion to store the most crucial information for SGD to recover, and then uses a Huffman coding to leverage the non-uniform distribution of the gradient scales. 
Our extensive experiments show that \projectname achieves a compression rate up to $28\times$ and recovery speedup up to $5.77\times$ over a state-of-the-art algorithm (SCAR).

\end{abstract}

\section{Introduction}
\label{Introduction}

Efficient construction of checkpoints (snapshots) has been increasingly important to deep learning research. In the arms race of developing more accurate models, researchers utilize heavier computing infrastructure and develop deeper and larger models. Without proper infrastructure support, the research process inevitably becomes fragile. For example, distributed computation fails from time to time, leading to the excessive need to re-train models~\cite{qiao2018fault}. Diagnosing deep learning models also evolves to a complex procedure partly because that the community has a better understanding of deep learning models and produces more rules for ``debugging'' them. Some common errors include gradient explosion~\cite{goodfellow2016deep}, ``divide by zero''~\cite{ioffe2015batch}, and dead activation. This calls for the need to construct ``breakpoints,''  resembling those used in debugging computer programs, so that researchers can conveniently jump to the state right before the model ``crashes'' in the training. 

Producing checkpoints frequently enables failed training process to restart with minimum wasted time, and serves as breakpoints for debugging models. So far the standard practice of constructing checkpoints is primitive. The most common practice is to save the model state directly, counting on that the backend system is sufficiently robust so that this operation does not become a bottleneck~\cite{baylor2017tfx}. Attempts of partially storing model states are also examined~\cite{qiao2018fault} but these works usually focus on recovery speed, instead of directly tackling system issues. 

The most pronounced technical challenge here is that deep models are usually large, so producing frequent checkpoints creates unmanageable burdens to both I/O and storage, even under modern distributed platforms \cite{abadi2016tensorflow, li2014scaling, low2012distributed}. Therefore, this leads to our question: 

{\quote{\textbf{Research Q:} \emph{How can we compress model checkpoints?}}}

We specifically aim to design a \emph{lossy} compressing scheme, addressing two criteria simultaneously. First, like standard compression problems, we need to maximize the compression rate. Second, the scheme needs to be optimized for the downstream application of \emph{training.} When a model restarts from our lossy checkpoints, it needs to efficiently resume to the most recent state (e.g., restart from a failed process or reach the state preceding the crash). 

Compression of model states is a new technical problem that requires addressing cross-cutting constraints from information theory, learning algorithm, and system design. We need to leverage statistical patterns encoded in the model state and factor in how the model states interact with a learning algorithm (more specifically, stochastic gradient type algorithms in the deep learning setting). This means neither standard lossy compression algorithms nor recently developed model compression algorithms~\cite{han2015deep, courbariaux2015binaryconnect, hong2016convergence, leng2018extremely, lin2016fixed} directly work in our setting. Standard lossy compression algorithms aim to minimize reconstruction error but our end goal is to enable a learning algorithm to ``quickly recover.'' Model compression techniques aim to transform a (static) model into a simpler one while ensuring the forecasts are not perturbed much whereas in our setting we need a reliable coding scheme that functions well throughout the entire dynamic process of learning, which is an orthogonal and perhaps more challenging goal. In addition, our algorithm must be efficient and scalable so that it can be executed frequently. 

\myparab{Our solution.} To achieve our aims, we focus on a delta-encoding scheme~\cite{mogul1997potential}, tracking only the information on the difference between two checkpoints. Under this scheme, we examine whether we can cut the least useful information (with respect to training) from the model state, and ensure that the remaining information is amenable for compression.  A perhaps surprising message here is that $\ell_2$-norm reconstruction error for the ``delta'' appears to be an ineffective metric for minimizing the recovery time. Instead, our algorithm first removes all the parameters with inconsequential updates, and then quantizes the remainder information. These strategies resemble those used in distributed training with the goal of minimizing communication cost~\cite{alistarh2017qsgd}. After we obtain the most significant information for portion of parameter updates, we represent them in suitable format and apply a Huffman coding to further compress these bits, so that the compression rate can be at the information theoretic limit. This strategy resembles recent techniques for model compression~\cite{han2015deep, wu2016quantized, park2017weighted, zhou2017incremental,  rastegari2016xnor}.

The contribution of this paper includes: 
\begin{itemize}
    \item Proposal of a fundamental research question on compressing model states for training recovery. 
    \item Characterization of a family of compression schemes that can efficiently track the learning process, based on a stylized model we develop. 
    \item Design of a lossy coding scheme with high-compression rate that integrates both classical compression techniques and recent ones developed for distributed learning and model compression. 
    \item Optimization of training systems that minimizes the overhead of producing checkpoints on the fly. 
\end{itemize}

Our extensive evaluation demonstrates that by simultaneously leveraging techniques from distributed training and model compression, our algorithm delivers a solution (called \projectname, LC refers to Lossy Compression) with a compression rate of {\textbf{up to 28x}} and superior recovering time---achieving up to $5.77\times$ recovery speedup over a state-of-the-art algorithm (SCAR). 



\section{Our approach}\label{sec:approach}

We now describe our compression framework. We introduce a stylized model for the learning process to facilitate the analysis of the system design trade-off. Then we explain our design principles, determined by both the stylized model and our extensive experiments. 

\myparab{Our model.} A ``high-dimensional'' vector $\mmu \in \reals^n$ represents the model state. An iterative algorithm (e.g., stochastic gradient descent) is used to gradually move the model state vector $\mmu$ toward a local optimal point $\mmu^*$. Let $\mmu_t$ be the model state at the $t$-th round. In our stylized model, we assume $\mmu_t$ performs a (drifted) random walk that converges to $\mmu^*$. Specifically, we use the following process to model $\mmu_i$'s trajectory. Let $L = \|\mmu_0 - \mmu^*\|.$ 
\begin{equation}\label{eq:model}
    \mmu_{t + 1} = \mmu^* + \eta (\mmu_{t} - \mmu^*) + \epsilon_t, 
\end{equation}
where $\eta$ and $L$ jointly model the convergence rate of the algorithm, and $\epsilon_t$ is a random noise component to reflect the stochastic nature of SGD. When $\eta$ is set to be a small constant, the model characterizes those algorithms that have linear convergence rate. When $\eta = (1-1/L)$, this model characterizes those algorithms whose convergence rates are $1-1/t$~\cite{boyd2004convex}. 
While our model does not captures the detail of many SGD algorithms, because different SGD algorithms have different convergence rate, designing a unifying model that highlights design trade-offs requires us to make simplifying assumptions.


\myparab{Our design principles.} We next describe our design principles.


\mypara{P1. Minimize irritation to SGD.} When we design lossy compression scheme, a portion of information is inevitably lost, causing performance degradation to a learning algorithm. We find that we should not simply use 
 $\ell_2$ reconstruction error to measure degradation of SGD. This can be best illustrated by the stylized model. For simplicity, let $\mmu^* = 0$, so $\mmu_{t + 1} = \mmu_{t} - 
\left((1 - \eta)\mmu_t + \epsilon_t\right)$. The delta term we want to compress is $\left((1 - \eta)\mmu_t + \epsilon_t\right)$. When we use a lossy compression, it corresponds to adding an additional noise term that is a function of $\mmu_t$ and $\epsilon_t$. So with the compression scheme, the new learning process becomes 
$\mmu_{t + 1} = \mmu_t - \left((1-\eta)\mmu_t + \epsilon_t + f(\mmu_t, \epsilon_t) \right)$. Observing that as long as $\E[f(\mmu_t, \epsilon_t) \mid \mmu_t, \epsilon_t] = 0$, and $\mathrm{Var}(f(\mmu_t, \epsilon_t) \mid \mmu_t, \epsilon_t)$ is dominated (smaller than) by $\mathrm{Var}(\epsilon_t)$, then the convergence quality remains unchanged, by standard results from stochastic approximation~\cite{lai2009martingales,kushner2003stochastic}. 

There are many constructs that satisfy the expectation and variance constraints. Let us consider an example of keeping the most significant bit of $\left((1-\eta)\mmu_t + \epsilon_t\right)$  by using standard randomized rounding~\cite{alistarh2017qsgd}. Because of the nature of the rounding algorithm, the expectation is $0$. In addition, because the most significant bit is kept, the information loss in rounding will not be greater than  $\|\left((1-\eta)\mmu_t + \epsilon_t\right)\|_2 = O(\mathrm{std}(\epsilon_t))$ under a mild assumption that $\epsilon_t$'s standard deviation also scales proportionally to $\|\mmu_t\|$ over time. 
Therefore, this rounding scheme does not affect the performance of the training algorithm. In general, the 1-bit encoding is a special case of quantization. A wide family of quantization schemes will satisfy the expectation and variance constraint. Our algorithm will explore this trade-off. 

Note also when we minimize $\ell_2$ reconstruction error, this corresponds to keeping top-$k$ heaviest entries in $\mmu_{t + 1} - \mmu_t$.

\begin{figure*}[t]
 \centering
  \includegraphics[width=0.95\textwidth]{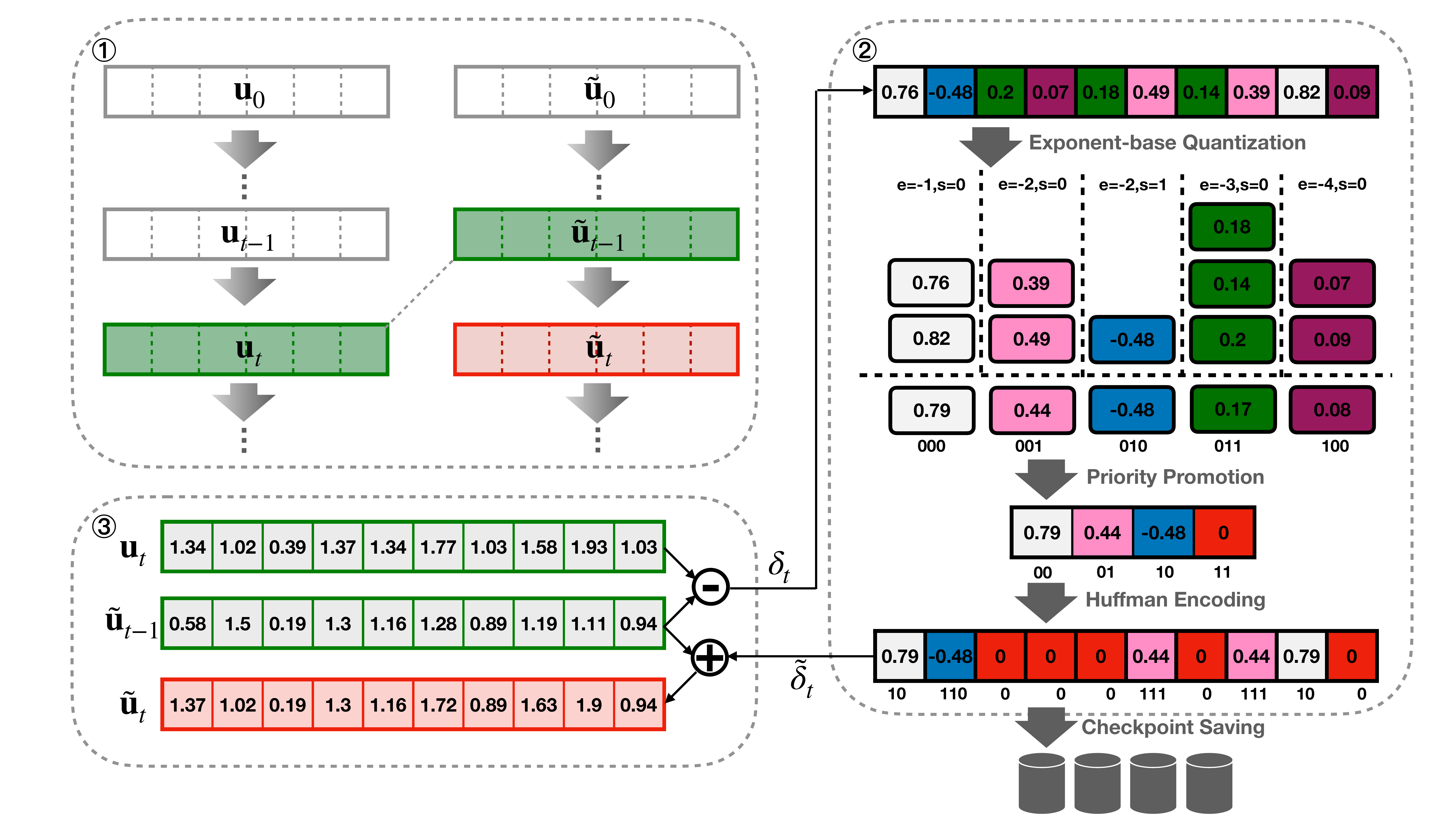}
   \caption{{\bf \projectname overview.}}
   \label{fig:overview}
\end{figure*}

\mypara{P2. Maximize redundancies in residual information.} Our compression scheme also needs to ensure the information we keep exhibits large redundancy, as measured by entropy. This will enable us to use traditional coding schemes such as Huffman code
to compress the data at the information theoretic limit. 

The interplay between P1 and P2 highlights the unique structure of our compression problem. This can be best illustrated by a compression scheme called TOPN. This compression scheme keeps the largest elements in $\delta_t$. We observe \emph{(i)} while this scheme minimizes $\ell_2$ reconstruction error, it \emph{does not} have superior recovery time. Many other compression schemes that possess the aforementioned properties recover equally fast, as suggested by our stylized model. \emph{(ii)} It is difficult to perform compression for the TOPN scheme. TOPN scheme usually needs to track 10\% of all the entries in $\delta_t$ to be effective. The overhead of tracking the \emph{locations} of these elements is surprisingly high. This is because in part that the vector is not sufficiently sparse so sparse matrix representation does not help. 

Our solution, on the other hand, carefully complies P1 and circumvents the need to track the locations of the entries we keep and thus achieves significantly higher compression rate. 


\mypara{P3. Do not use random projections and/or sketches.} Notably, we discover that sketch-based randomized projection techniques (e.g.,~\citet{woodruff2014sketching}) \emph{harm} the compression. Roughly speaking, sketches compress information by projecting multiple numbers into one cell. While this could speed up query time, it only irritates the gradient descent algorithm in our setting. Consider a toy example in which $\mmu_t \in \reals^2$ and the optimal point $\mmu^* = (0, 10)$. Let $\mmu_t = (5, 5)$ be the current state so the gradient is along the direction $(-1, 1)$. When we apply sketches (say CountMin sketches), it collapses the direction $(-1, 1)$ into a single point $0$. When we make a query, the gradients for both coordinates are incorrect. Sketches are more useful when the entries in the gradient vector are heterogeneous and queries need to be answered at ``line rate'' (e.g., do not slow down the training ~\citet{ivkin2019communication}). Here, when a model needs to be recovered from a checkpoint, the job is less time-sensitive. Therefore, even we face heterogeneous parameters, it is more effective to carefully disentangle crucial information from inconsequential ones than using arbitrary random projections.

\section{\projectname-based SGD}
\label{Design}

\begin{algorithm}[t]
    \caption{LC-CHECKPOINT-BASED SGD}
    \label{alg:check-point-sdg}
    {\bfseries Input:} $\mmu^*$, $\mmu_0$, $\eta$
    \begin{algorithmic}[1]
    \State Initialize $\tilde \mmu_0 = \mmu_0$.
    \For{$t=1$ {\bfseries to} $T$}
    \State Update model state: $\mmu_{t} = \mmu^* + \eta (\mmu_{t-1} - \mmu^*) + \epsilon$
    \State Compute distance: $\delta_{t} = \mmu_{t} - \tilde \mmu_{t-1}$ 
    \State Quantize ${\delta_t}$: $\tilde \delta_t =  \textproc{QUANTIZE}({\delta_t})$
    \State Compress $\tilde \delta_t$ by Huffman coding and save to disk
    \State Update checkpoint state: $\tilde \mmu_{t} = \tilde \mmu_{t-1} + \tilde \delta_t$
    \EndFor
    \end{algorithmic}
    {\bfseries Output:} $\mmu_{T}$, $\{\tilde \delta_t \mid t \in [T]\}$
\end{algorithm}

We now describe our solution \projectname (LC refers to Lossy Compression). 
See Figure~\ref{fig:overview} for a working example and Algorithm~\ref{alg:check-point-sdg} for a workflow. For simplicity, we assume that our system maintains a checkpoint $\tilde \delta_t$ for each iteration. We slightly abuse $\delta_t$ to refer to both the compressed data and the real vector it represents. It is straightforward to downsample our operations to construct a checkpoint every $k$-iterations. 
Our solution consists of two major components. 

\myparab{C1. Approximate tracking by delta-coding.} At each step, our system maintains an approximation $\tilde \mmu_t$ of the ground-truth state. We simply set $\tilde \mmu_t = \mmu_0 +  \sum_{i \leq t}\tilde \delta_i$, where $\mmu_0$ is the initial state of the model. Our system continuously maintains and updates $\tilde \mmu_t$ at the background (line 7 in Algorithm~\ref{alg:check-point-sdg}). Our major compression task is to properly track the ``delta'' between the approximate state and ground-truth. Specifically, the compression task for the $t$-th iteration is $\delta_t = \mmu_t - \tilde \mmu_{t - 1}$. See \textcircled{3} in Figure~\ref{fig:overview}. 

\myparab{C2. Quantization and Huffman coding.} This component compresses $\delta_t$ through two steps, \emph{Step 1. Two-stage quantization.} We first perform an exponent-based quantization, and then a priority promotion operation. This operation intelligently drops inconsequential information between two consecutive states. \emph{Step 2. Lossless compression by Huffman.} Finally, the quantized distance vector is further compressed using Huffman coding. 

One can see that to reconstruct the model state at iteration $t$ from the checkpoints, we may simply compute $\mmu_t = \mmu_0 + \sum_{i = 1}^t \tilde \delta_t$. 

In what follows, Section~\ref{sec:QuanHuff} discusses C2 and Section~\ref{sec:opt} discusses additional system-level optimizations.

\begin{figure*} 
    \centering    
    \subfigure[Exponent distribution of $\delta$.] {
        \label{fig:distribution:a}     
        \includegraphics[width=0.45\linewidth]{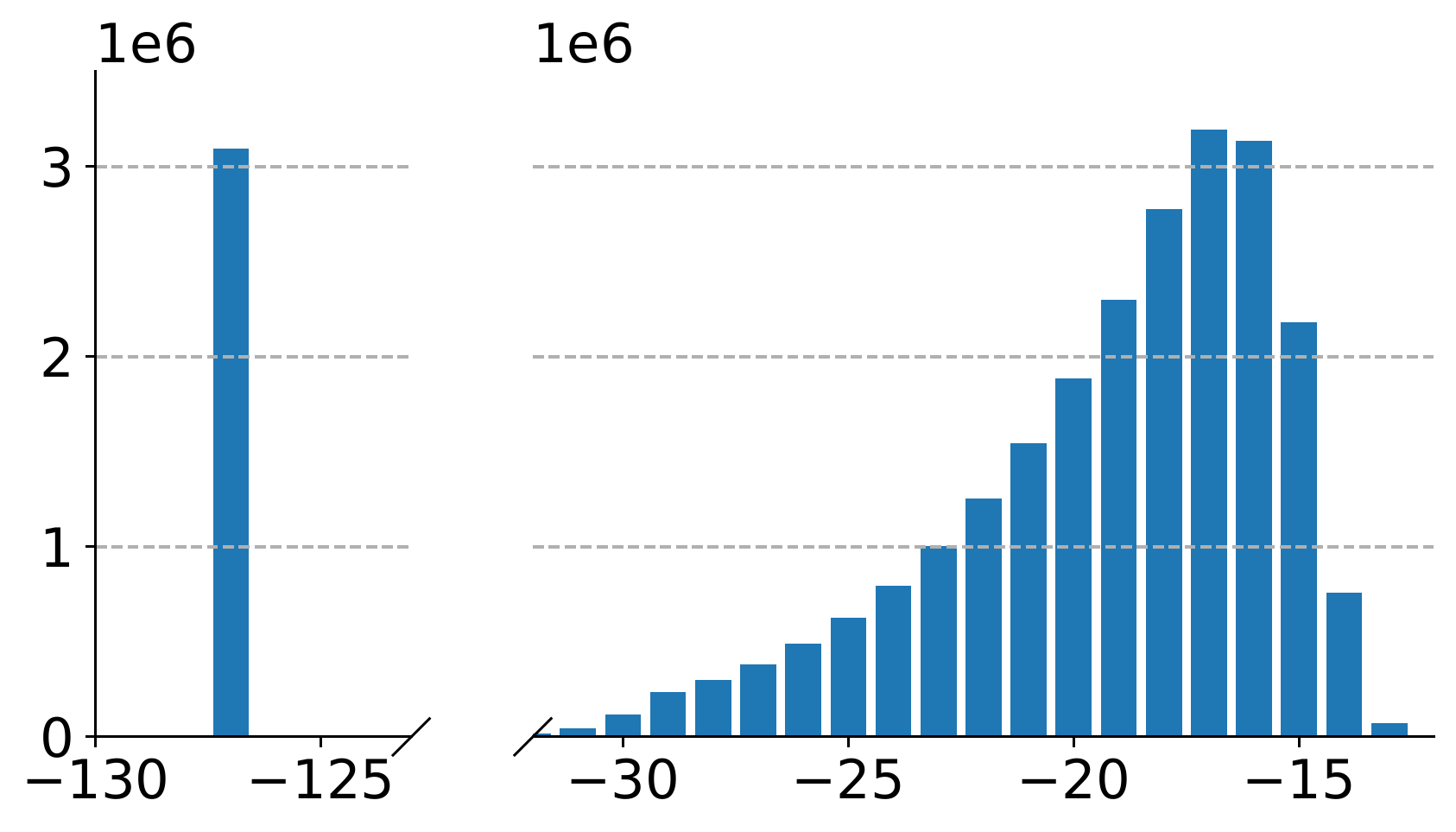}
    }     
    \subfigure[Exponent distribution of $\tilde \delta$ (3-bit promotion).] { 
        \label{fig:distribution:b}     
        \includegraphics[width=0.35\linewidth]{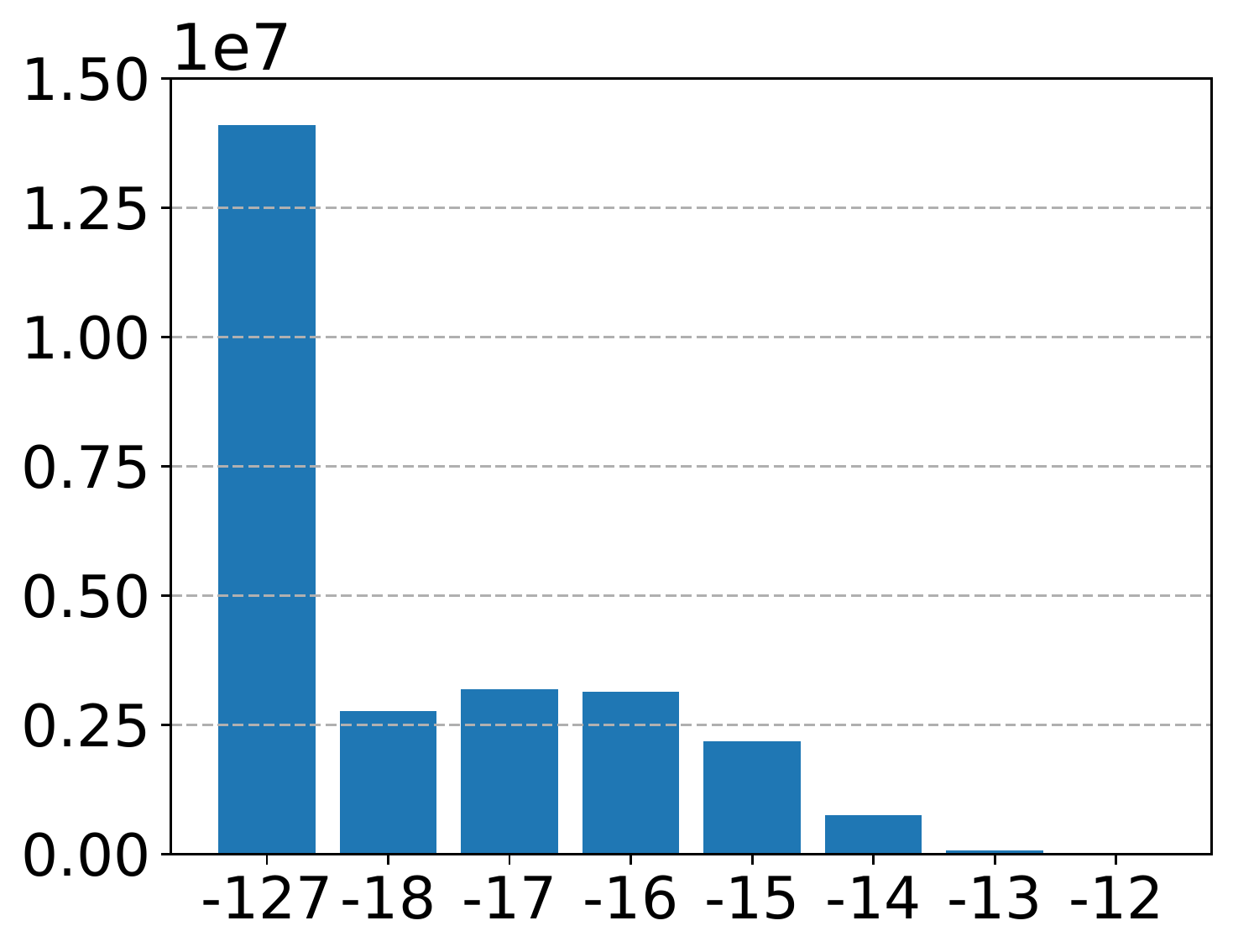}     
    }  
\caption{{\bf The distribution of all elements' exponent parts in the last convolutional layer of AlexNet.} When $e$ equals $-127$, the element value is $0$. The x-axis denotes the exponent part value, and the y-axis indicates the count of elements with this value.}     
\label{fig:distribution}     
\end{figure*}

\subsection{Quantization and Huffman coding}\label{sec:QuanHuff}

\subsubsection{Two-stage Quantization}
\label{section:quantization}
\projectname employs a novel two-stage pipeline to quantize $\delta_{t}$, which consists of two main sub-steps: exponent-based quantization and priority promotion.

\myparab{Exponent-based Quantization.} Recall that a floating point $v$ is represented by $v = (-1)^s \times m \times 2^e$, where  $s$ is the sign, $m$ is the mantissa, and $e$ is the exponent. Recall that $\delta_t = \mmu_t - \tilde \mmu_{t-1} \in \reals^n$ is a high-dimensional vector we aim to encode.  
Our exponent-based quantization works as follows: first, it partitions entries in $\delta$ into multiple buckets according to $e$ and $s$, i.e., it assigns the elements with identical exponents and signs to the same bucket. 
Our crucial observation from extensive experiments is that entries in $\mmu_t$ usually drift towards the same direction, so $\delta_t$ typically have the same sign. 
Next, our algorithm represents each bucket by the average of maximum and minimum values in the bucket. 

Figure~\ref{fig:overview} \textcircled{2} shows an example, in which, $\delta_t$ is quantized into five buckets (marked with five different colors). All entries in each bucket are then represented by a unique value. 

Indexing $k$ buckets requires $\log_2k$ bits. Because $\delta_t$ consists of $n$ floating points, each of which uses $b$ (e.g., $b \in \{32, 64\}$) bits, the compression rate is 
$r = \frac{nb}{n\log_2k+kb}.$

For example, in Figure~\ref{fig:overview}, $\delta$ has 10 elements (i.e., $n = 10$), each of which is represented by a single-precision floating point (i.e., $b = 32$). Thus, the original $\delta$ has $nb$, i.e., $320$ bits in total. Exponent-based quantization uses $5$ buckets (i.e., $k=5$). Thus, after quantization, $\delta$ has ($10 \times \log 5 + 5 \times 32 = 190$) bits. Therefore, the compressing rate ($r$) is $1.68$ (i.e., $320/190$).

It is critical to control the number of buckets $k$ to achieve an optimal compression ratio. Fortunately, the exponent-based bucketing can control $k \leq 2^9$ for single-precision floating point elements, and control $k \leq 2^{12}$ for double-precision. \footnote{Single-precision floating point numbers use 8 bits to store $e$, and together with a sign bit---that is why $k \leq 2^9$. Similarly, double-precision numbers use 11 bits to store $e$.}
Our evaluation results (Section~\ref{section:size}) confirm that usually $k < 2^5$ suffices. Figure~\ref{fig:distribution:a} plots the distribution of all elements' exponent parts in the last convolutional layer of AlexNet.

\myparab{Priority Promotion.} 
We further improve the compression ratio by limiting the number of buckets with a priority promotion approach.  Our crucial observation is that when $\delta_{t, i}$ is excessively close to $0$ (i.e., $\tilde \mmu_{i, t - 1}$ is close $\mmu_{i, t}$), it is more effective to batch the updates (i.e., do not update the $i$-th entry of $\delta_t$ until it becomes substantial). Note also this is conceptually different from minimizing construction errors. Minimizing construction errors corresponds to exactly keeping track of the heaviest entries in $\delta_t$, whereas we both remove excessively small entries and quantize large entries (as done in the previous step).  Specifically, we propose $x$-bit priority promotion. It keeps $2^x-1$ buckets with larger $e$ only and merges the rest buckets into one with a unique value of $0$. In other words, priority promotion  updates $\tilde \mmw_i$ with a larger distance to $\mmw_i$ with a higher priority. It limits the index of buckets within $x$ bits.

Figure~\ref{fig:overview} (Priority Promotion) uses 2-bit priority promotion to control the number of buckets under 4. It merges the green and purple buckets into a red one that is represented by a value $0$. Indexing these buckets only needs $2$ bits. Figure~\ref{fig:distribution:b} gives a real example of $3$-bit priority promotion for the last convolutional layer in AlexNet.

\subsubsection{Huffman Coding}
\label{section:huffman}

Finally, observing the number of elements in each bucket is highly non-uniform in most learning processes, we use Huffman coding~\cite{van1976construction} to further compress the bucket. 
For example, Figure~\ref{fig:distribution:a} plots the distribution of all elements' exponent parts in the last convolutional layer of AlexNet. This distribution shows a skewed behavior, 

thus more suitable for Huffman coding. Our crucial observation is that priority promotion \emph{further aggravates} the skewness of this distribution (Figure~\ref{fig:distribution:b}), thus marrying quantization with Huffman coding produces more than ``sum of parts'' benefits. Our later evaluation validates it (Section~\ref{section:size}).



\subsection{System Optimizations}\label{sec:opt}

\projectname also comprises several novel system-level optimizations as follows: 

\begin{itemize}[leftmargin=*,noitemsep,nolistsep]

\item {\bf Asynchronous Execution:} Because only the first step of \projectname depends on the model state, the rest steps can run simultaneously with the next iteration of SGD computation. This asynchronous (non-blocking) execution significantly reduces the checkpoint overhead, and mitigates the blocking of model execution. 

\item {\bf Checkpoint Merging:} To further reduce the recovery time, \projectname employs a helper process to merge multiple checkpoints into {\em super-step} ones, periodically. In case of any system crash, \projectname uses these {\em super-step} checkpoints for recovery. 

\item {\bf Huffman Code Table Caching:} 

The number of buckets may stay the same from one iteration to another, specifically after priority promotion. Thus, it is possible to reuse the Huffman code table (with only a simple sort of buckets according to the number of entries in each bucket) among different iterations without any rebuilding. \projectname comprises a lightweight cache to store the Huffman code table for each buckets count. 

\end{itemize}

\begin{figure*} 
    \centering    
    \subfigure[MLR on MNIST.] {
        \label{fig:e:overall:a}     
        \includegraphics[width=0.235\linewidth]{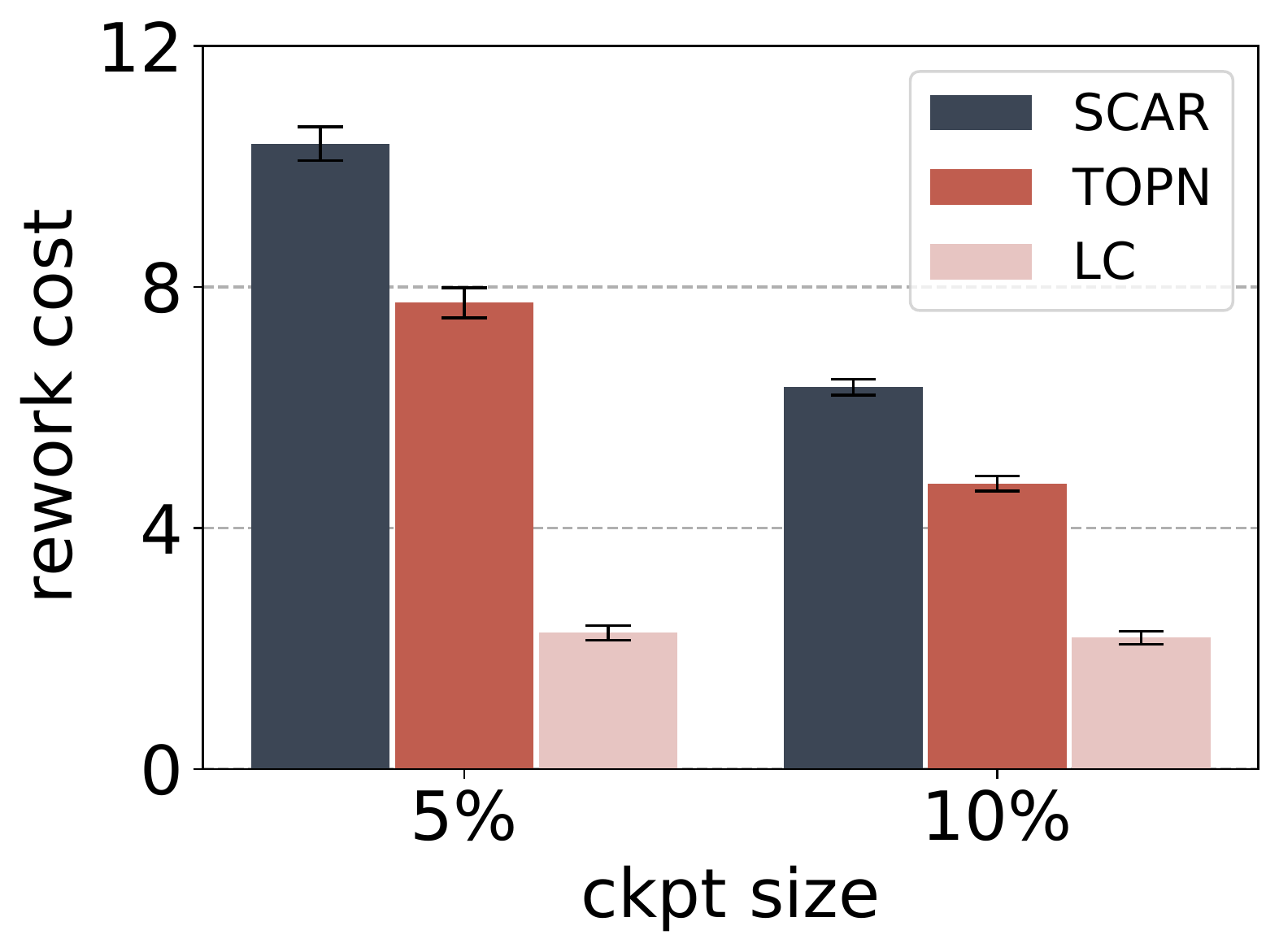}
    }     
    \subfigure[LeNet on MNIST.] { 
        \label{fig:e:overall:b}     
        \includegraphics[width=0.235\linewidth]{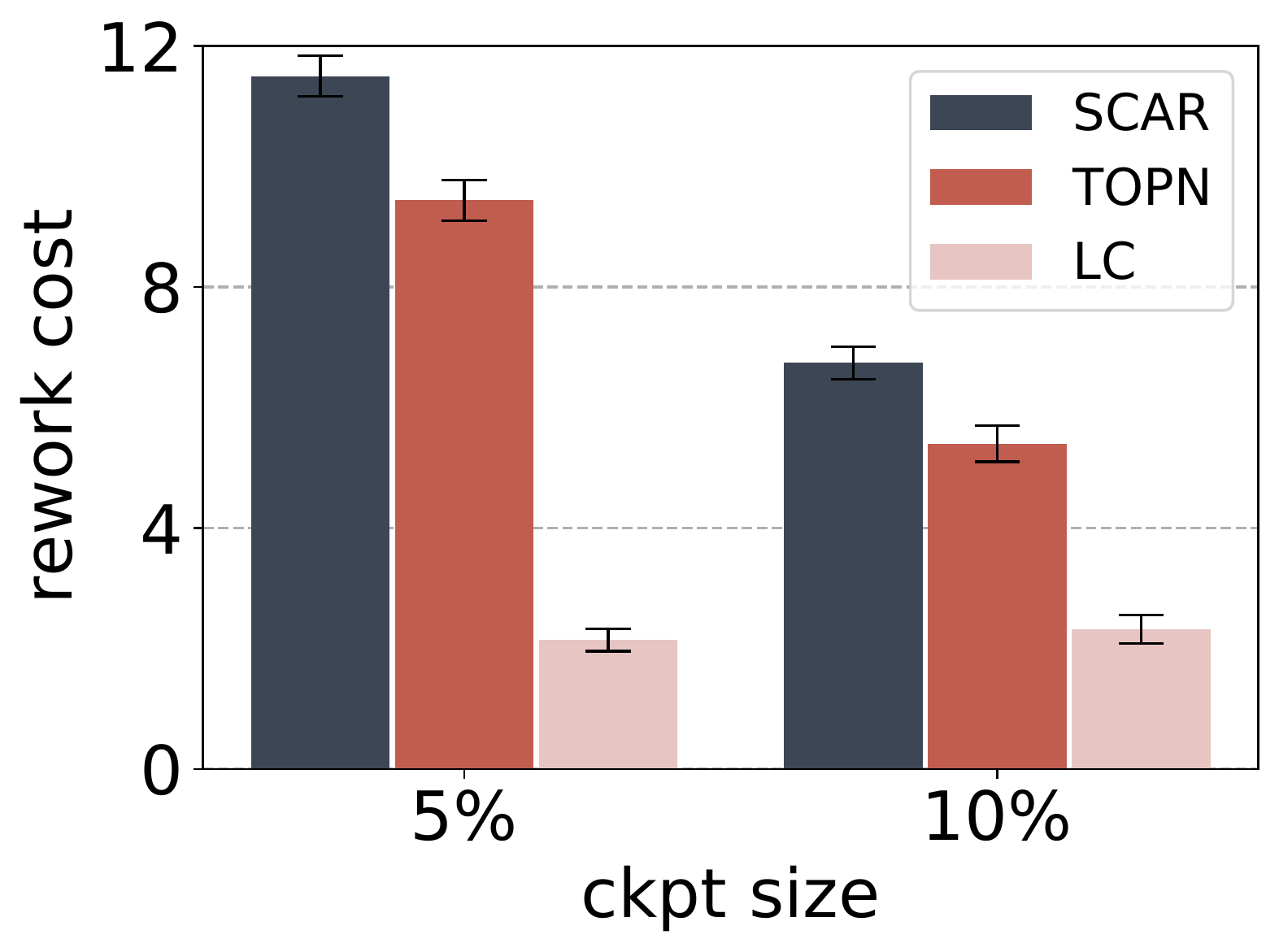}
    }    
    \subfigure[AlexNet on MNIST.] { 
        \label{fig:e:overall:c}     
        \includegraphics[width=0.235\linewidth]{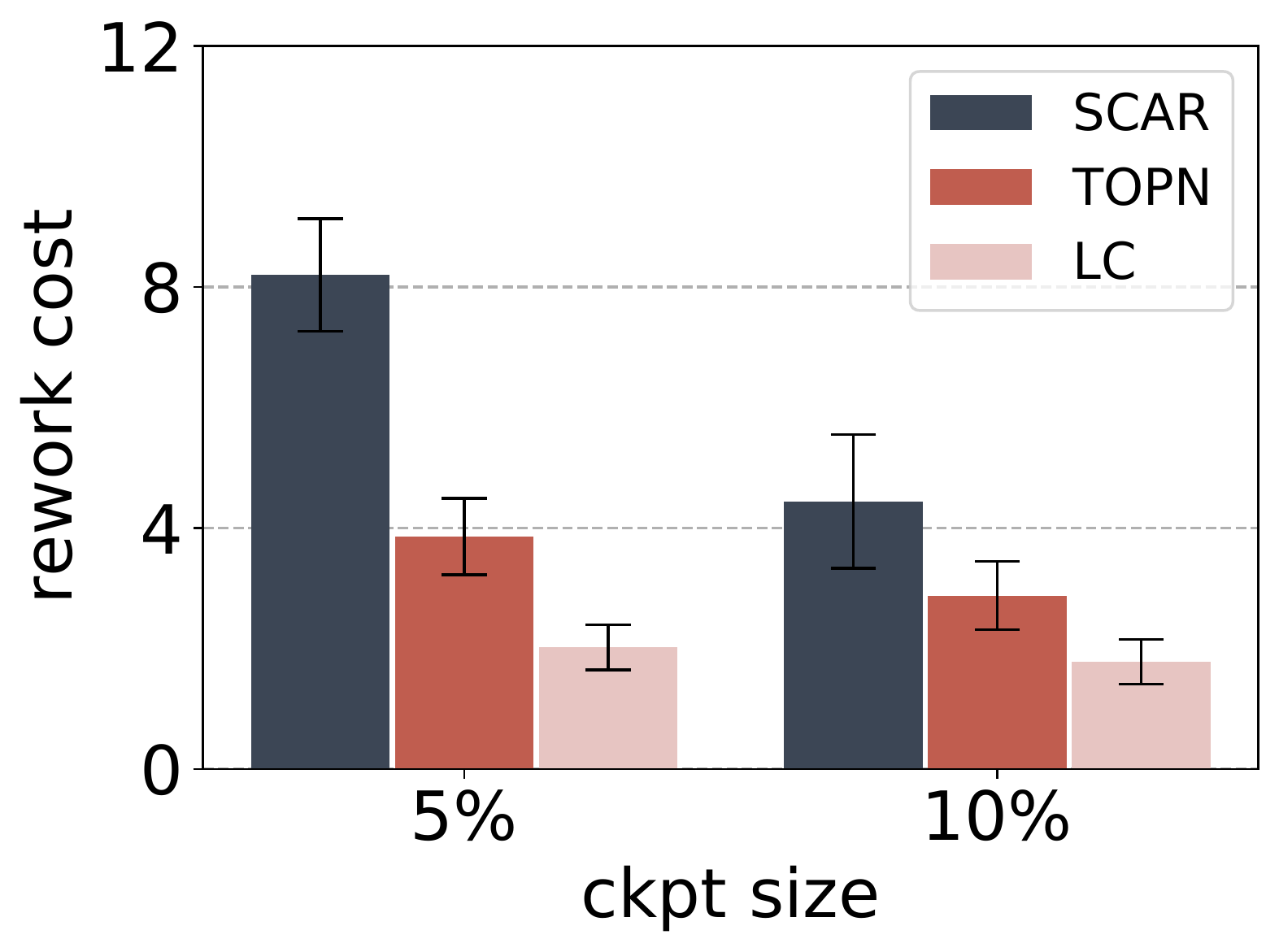}
    }
    \subfigure[MF on MovieLens.] { 
        \label{fig:e:overall:d}     
        \includegraphics[width=0.235\linewidth]{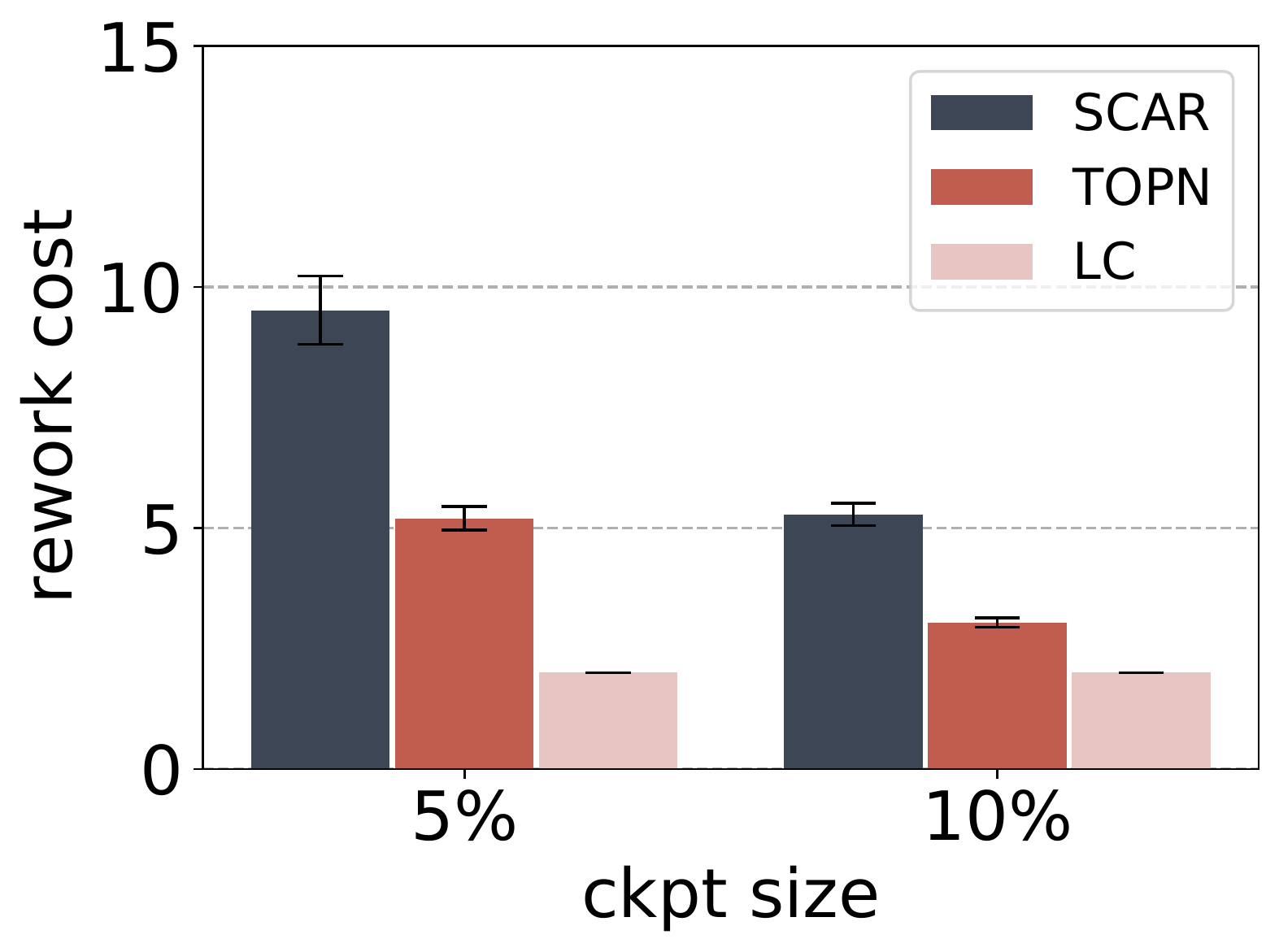}
    }
    
    \subfigure[MLR on FashionMNIST.] {
        \label{fig:e:overall:e}     
        \includegraphics[width=0.235\linewidth]{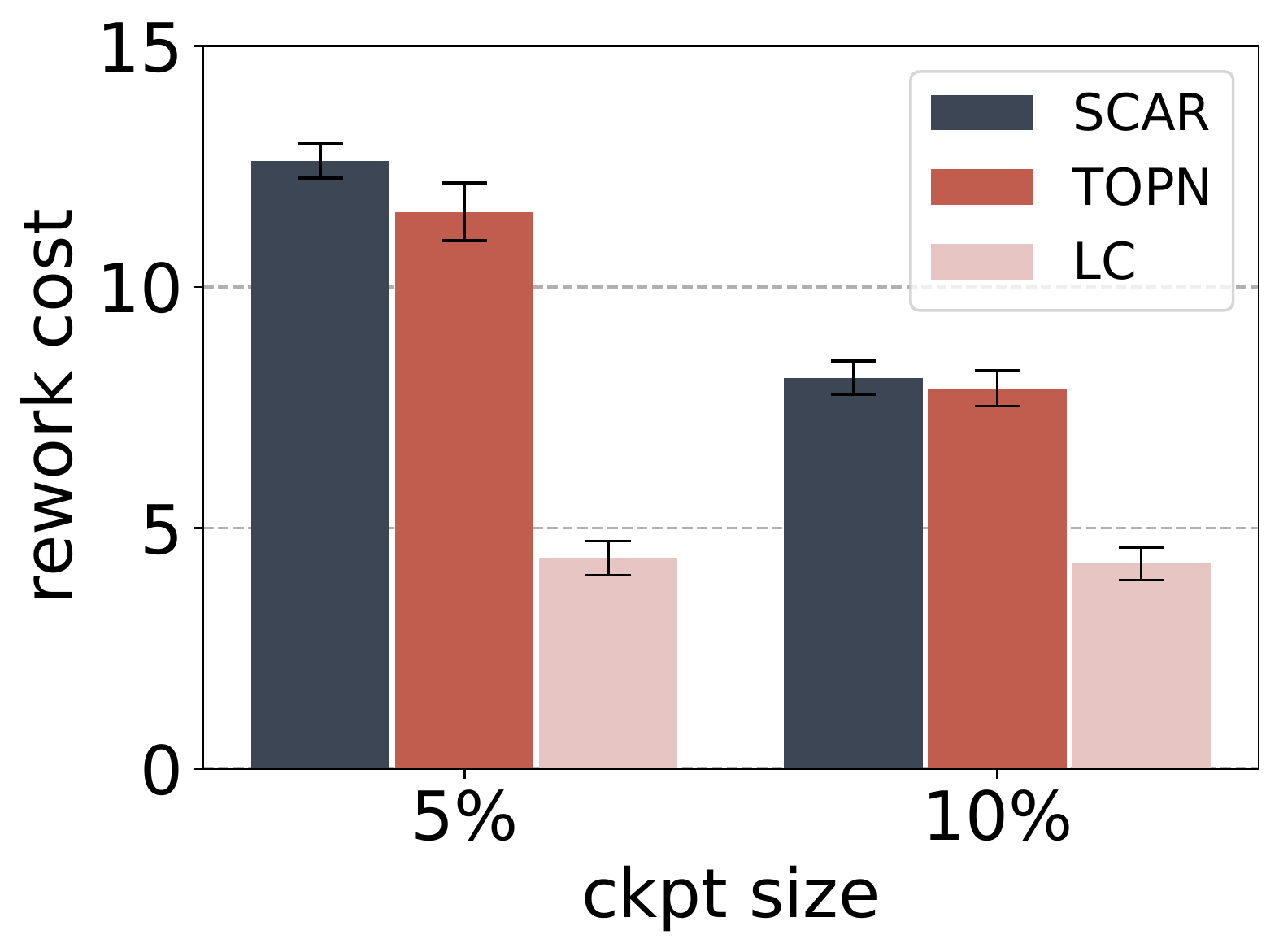}
    }     
    \subfigure[LeNet on FashionMNIST.] { 
        \label{fig:e:overall:f}     
        \includegraphics[width=0.235\linewidth]{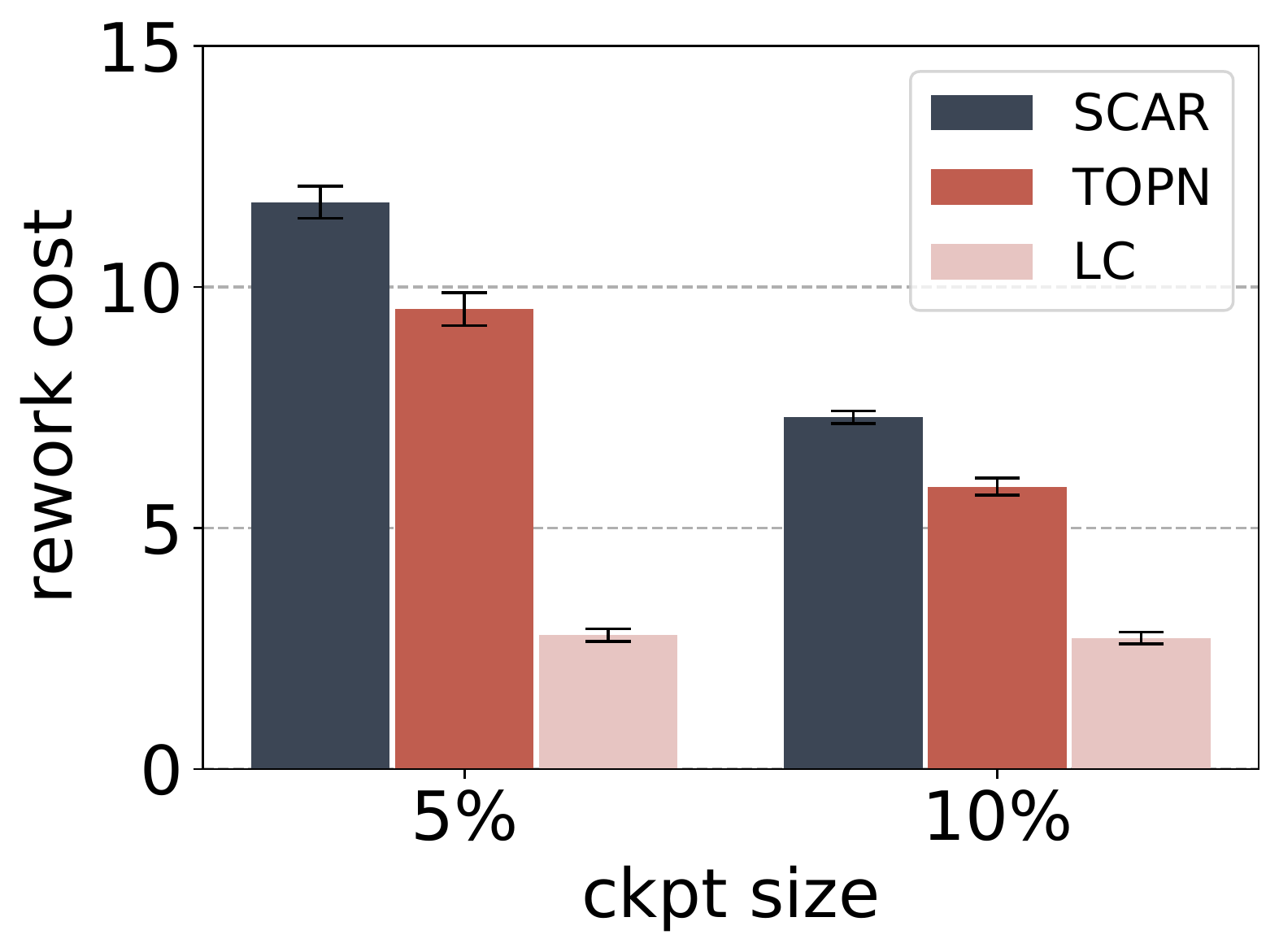}
    }    
    \subfigure[AlexNet on FashionMNIST.] { 
        \label{fig:e:overall:g}     
        \includegraphics[width=0.235\linewidth]{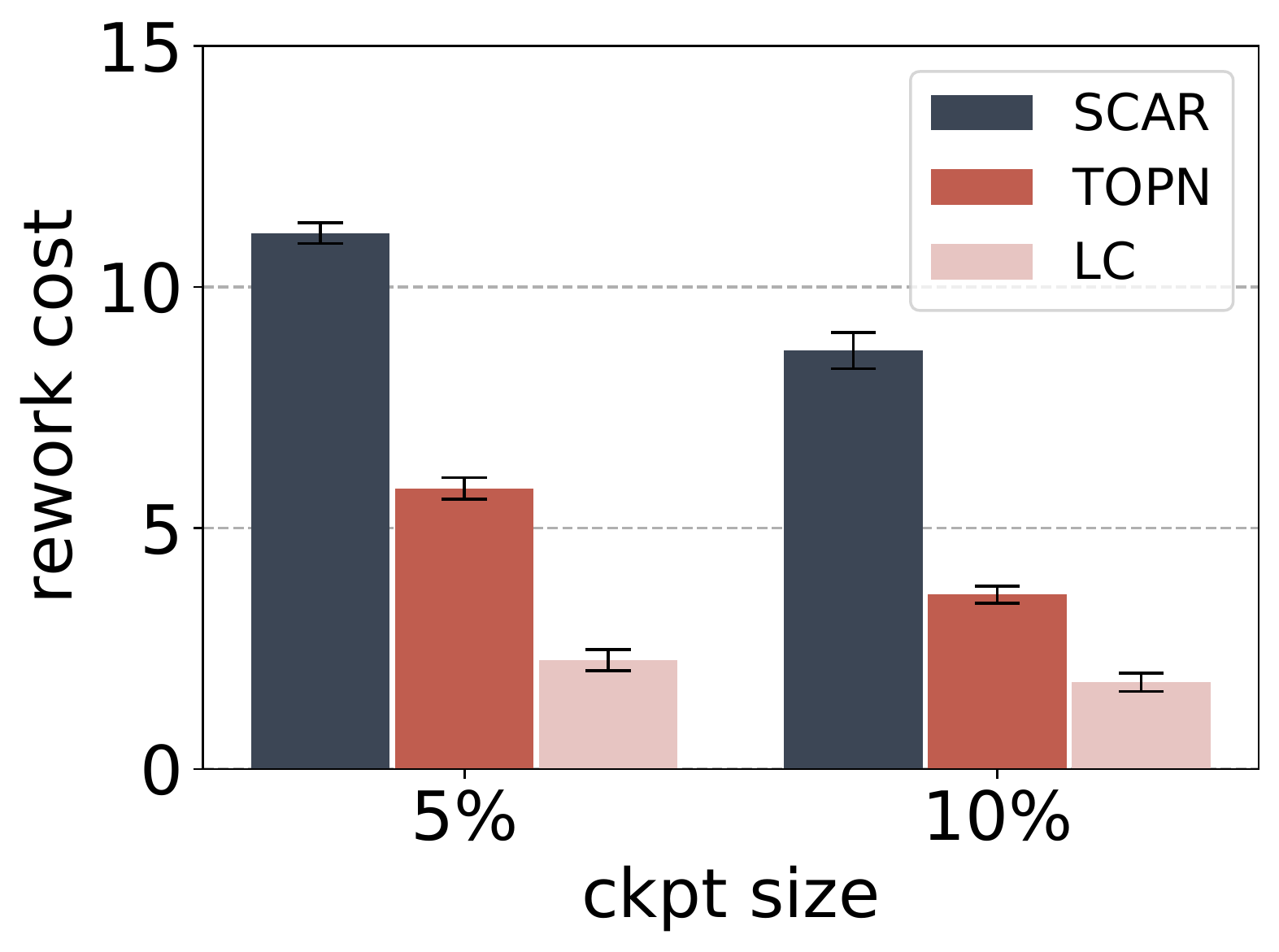}
    }
    \subfigure[MF on Jester.] { 
        \label{fig:e:overall:h}     
        \includegraphics[width=0.235\linewidth]{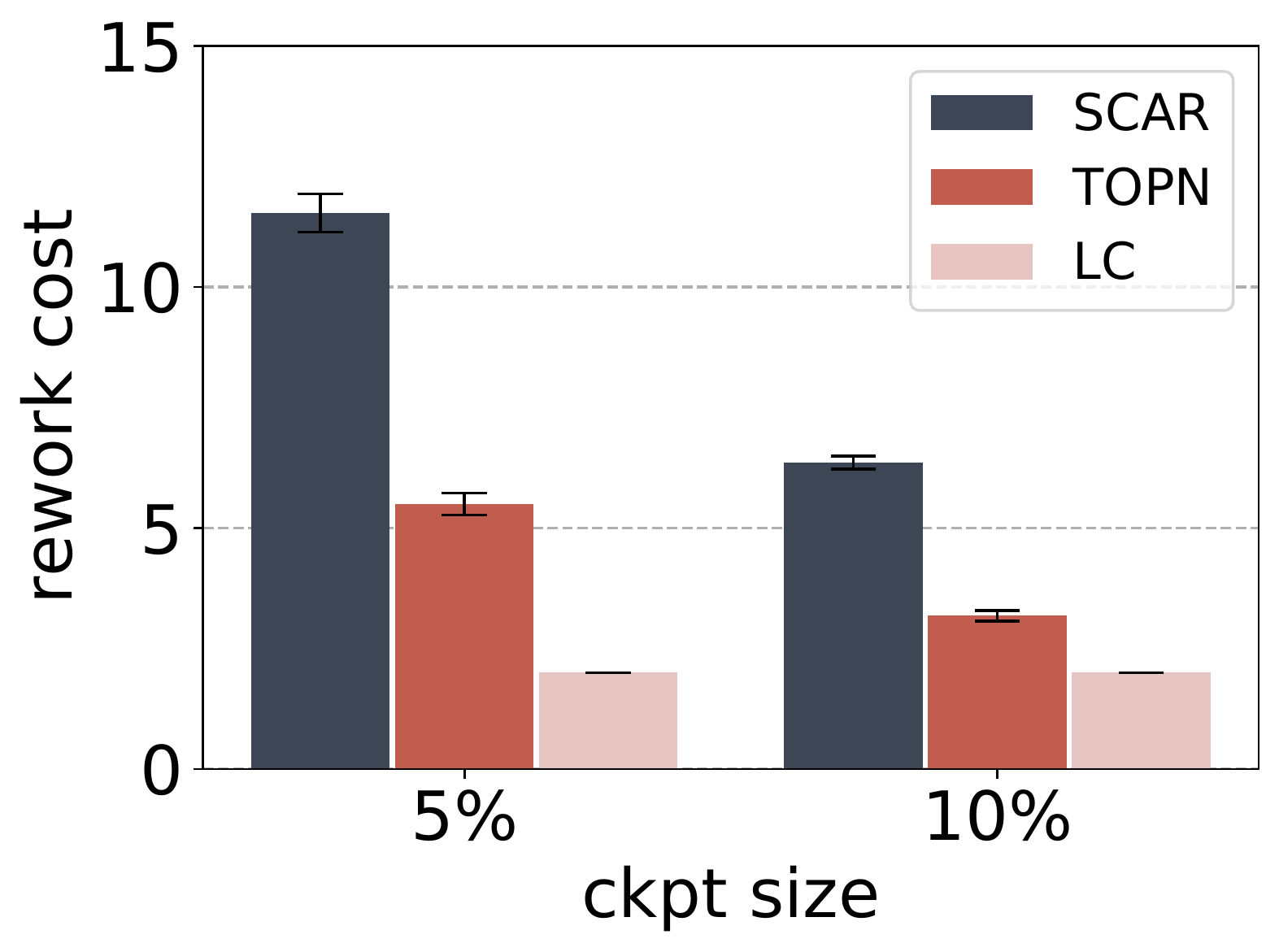}
    }
\caption{{\bf Rework cost comparison among \projectname, SCAR, and TOPN.} The x-axis indicates the ratio of the compressed checkpoint size over the full checkpoint size. The y-axis shows the rework iterations. The error bars indicate 95\% confidence intervals, calculated by repeating each trial 50 times. \vspace{-.2cm}}     
\label{fig:e:overall}     
\end{figure*}
\section{Experiments}
\label{experiments}

This section evaluates \projectname on four typical ML applications with three benchmark datasets, and compares it with previous efforts (SCAR~\citet{qiao2018fault} and a TOPN mechanism as mentioned in Section~\ref{sec:approach}) on recovery (rework) cost, compression ratio, and execution overhead, demonstrating the superiority of \projectname.

\subsection{Methodology}

\noindent{\bf Evaluation Objective:} This evaluation has four main objectives: (1) comparing \projectname' recovery (rework) cost with previous work; (2) evaluating the compression benefits brought by different approaches mentioned before; (3) specifically, validating the effectiveness of priority promotion; (4) confirming that \projectname incurs low overhead by an experiment case study. Our work is mainly compared with two state-of-the-art efforts: SCAR~\cite{qiao2018fault} and a TOPN mechanism. SCAR partitions the parameters and updates one partition in each iteration to reduce the checkpoint size. The TOPN mechanism only updates the parameters with the top-n largest distances to the previous iteration. The TOPN checkpoint is stored in a compressed sparse row (CSR) format.

\noindent{\bf ML Applications and Datasets:} \projectname is evaluated on four typical ML applications: Multinomial Logistic Regression ({\tt MLR}), LeNet-5 ({\tt Lenet})~\cite{lecun1998gradient}, {\tt AlexNet}~\cite{krizhevsky2012imagenet} and  Matrix Factorization ({\tt MF}). The first three applications are trained on {\tt MNIST}~\cite{lecun1998gradient} and {\tt FashionMNIST}~\cite{xiao2017fashion} datasets. The last one, {\tt MF} is trained on {\tt Jester}~\cite{goldberg2001eigentaste} and {\tt MovieLens10M}~\cite{harper2015movielens}.

\noindent{\bf Platforms and Evaluation Configurations:} Our experiments are conducted on a multi-core server with an Intel Xeon Gold 6138 Skylake CPU with 40 cores, each running at 2.0 GHz, and 192 GB DDR4 memory. The training is performed on a Tesla P100 GPU with 16GB High-bandwidth Memory (HBM).

\begin{figure*} 
    \centering    
    \subfigure[MLR on MNIST.] {
        \label{fig:e:size:a}     
        \includegraphics[width=0.235\linewidth]{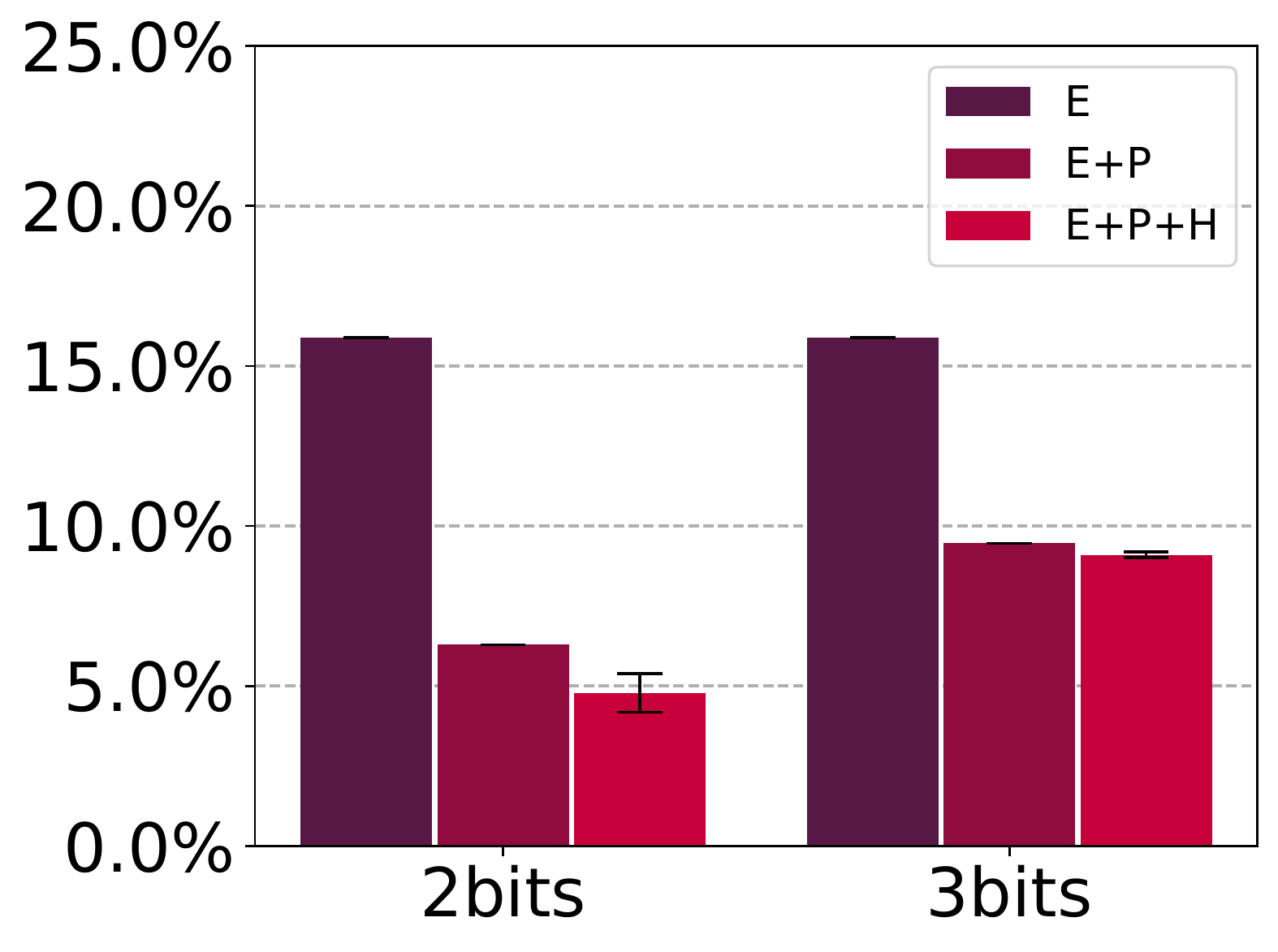}
    }     
    \subfigure[LeNet on MNIST.] { 
        \label{fig:e:size:b}     
        \includegraphics[width=0.235\linewidth]{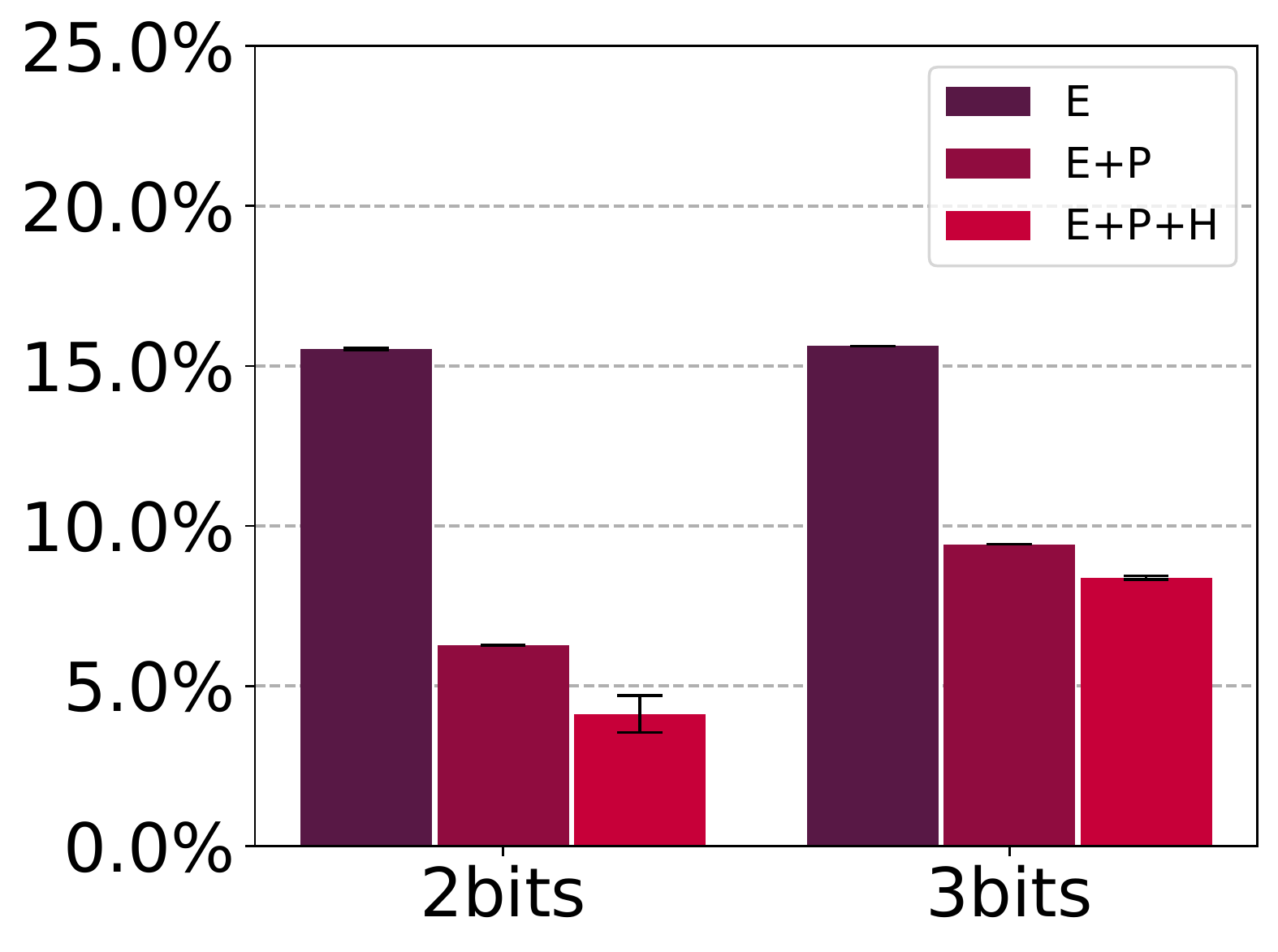}
    }    
    \subfigure[AlexNet on MNIST.] { 
        \label{fig:e:size:c}     
        \includegraphics[width=0.235\linewidth]{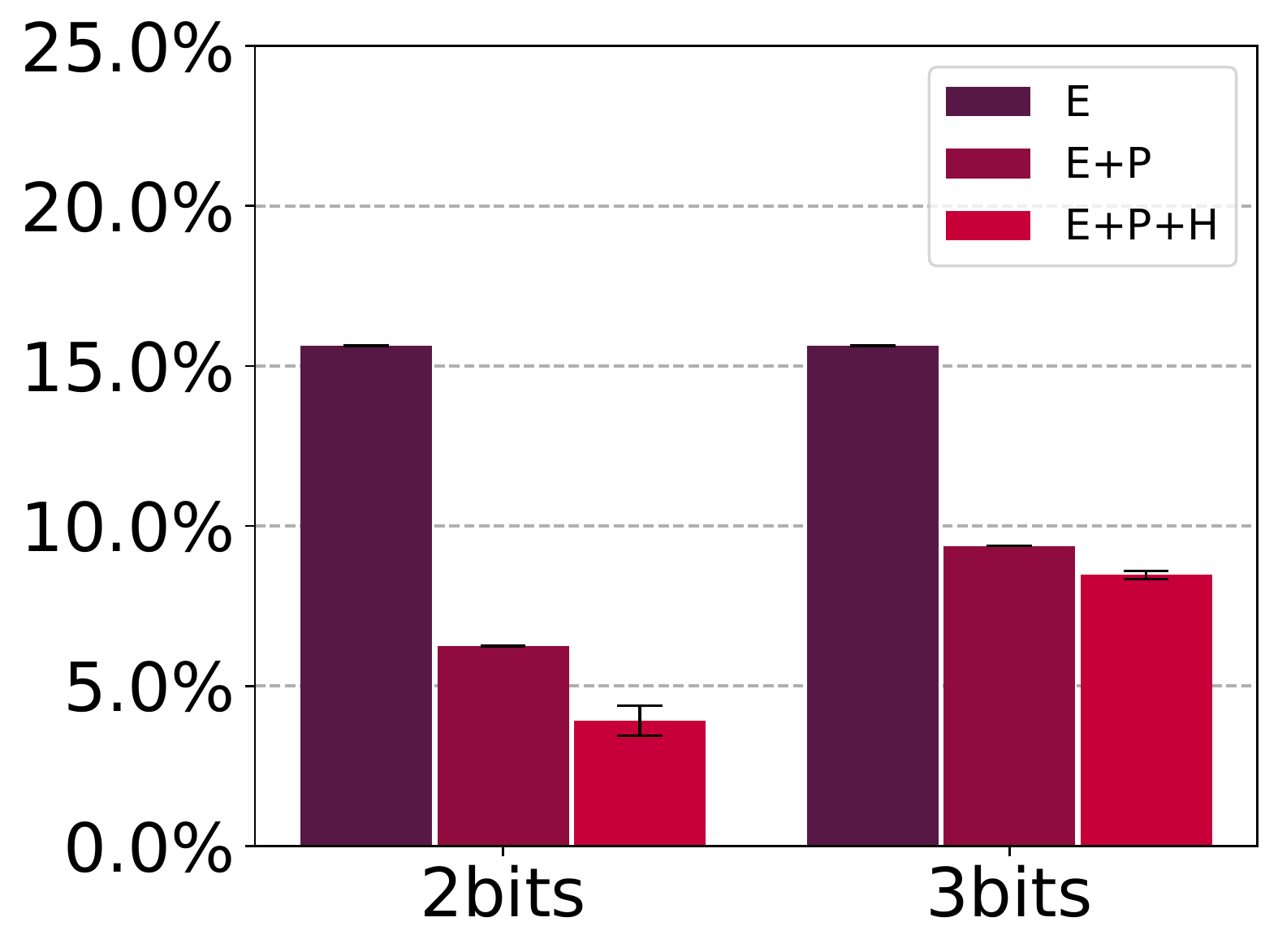}
    }
    \subfigure[MF on MovieLens.] { 
        \label{fig:e:size:d}     
        \includegraphics[width=0.235\linewidth]{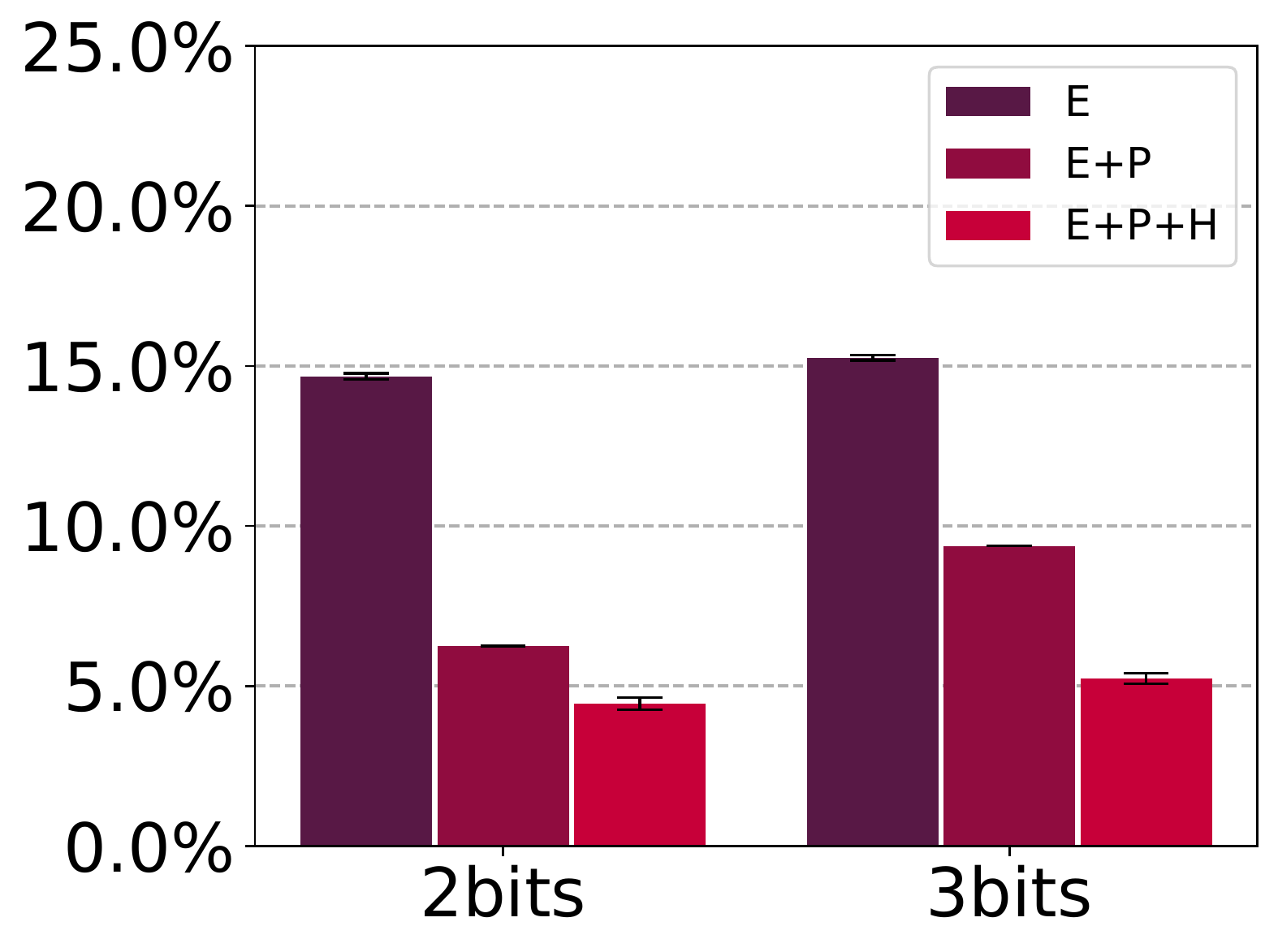}
    }
    
    \subfigure[MLR on Fashion-MNIST.] {
        \label{fig:e:size:e}     
        \includegraphics[width=0.235\linewidth]{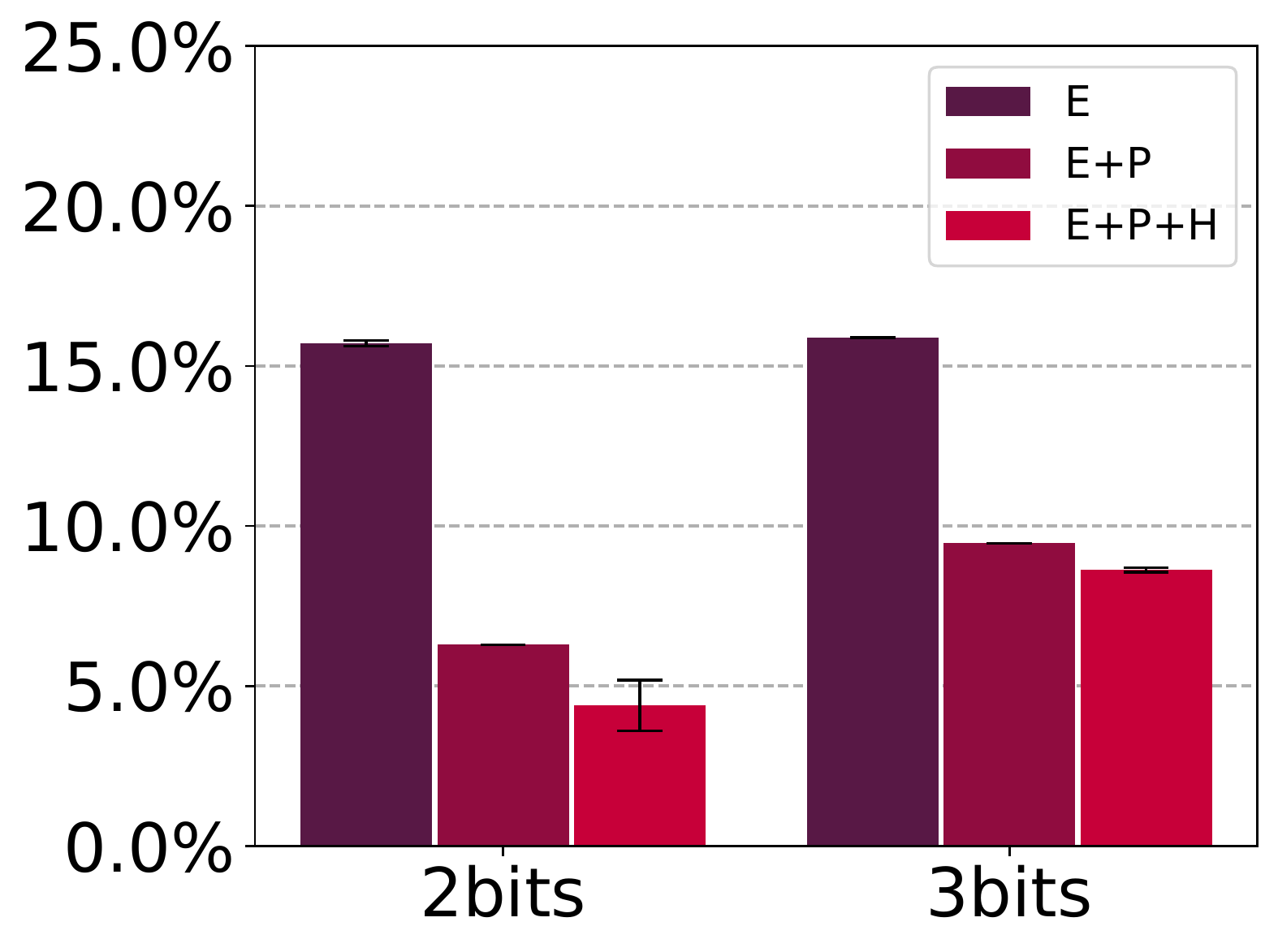}
    }     
    \subfigure[LeNet on Fashion-MNIST.] { 
        \label{fig:e:size:f}     
        \includegraphics[width=0.235\linewidth]{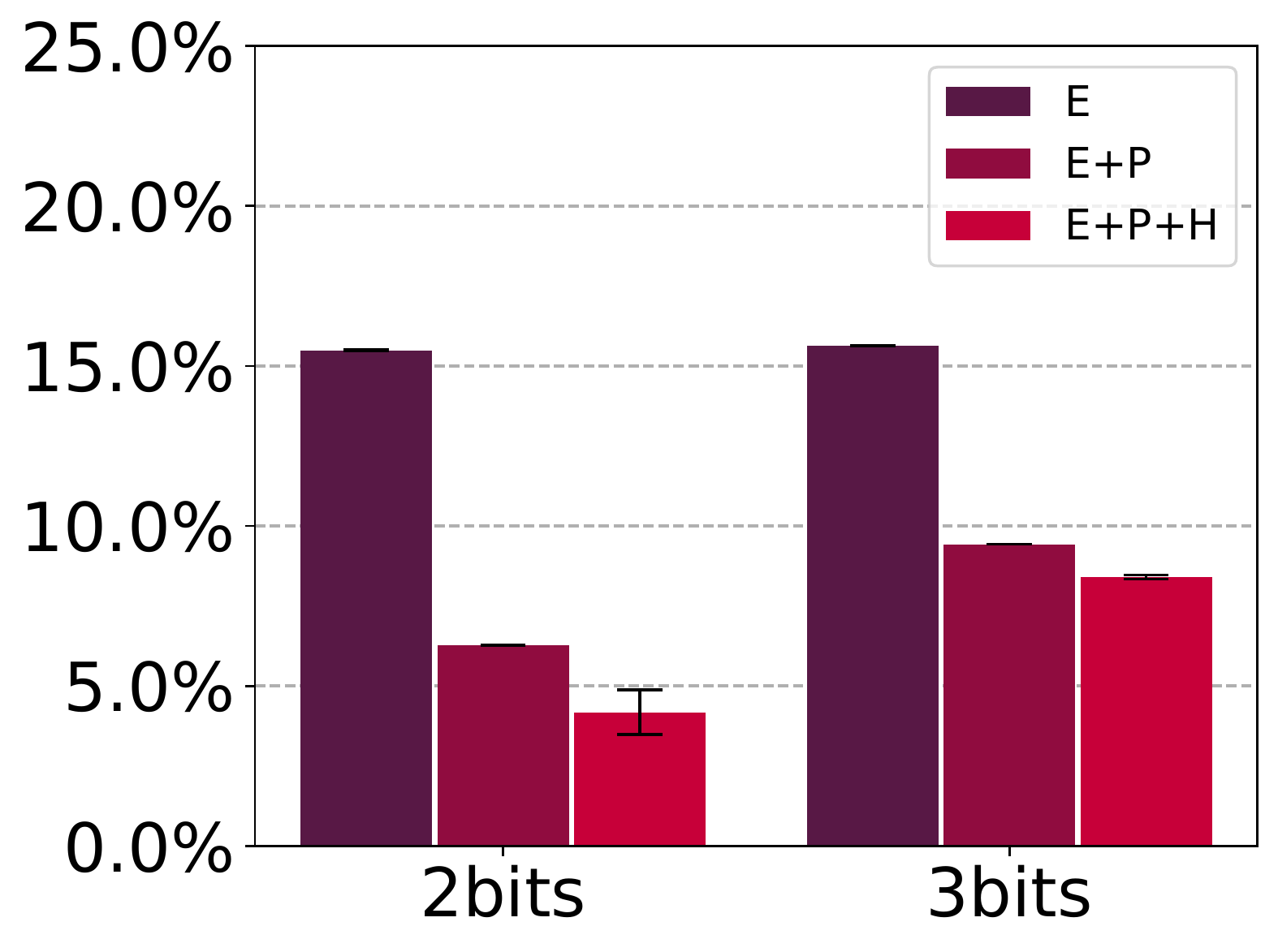}
    }    
    \subfigure[AlexNet on Fashion-MNIST.] { 
        \label{fig:e:size:g}     
        \includegraphics[width=0.235\linewidth]{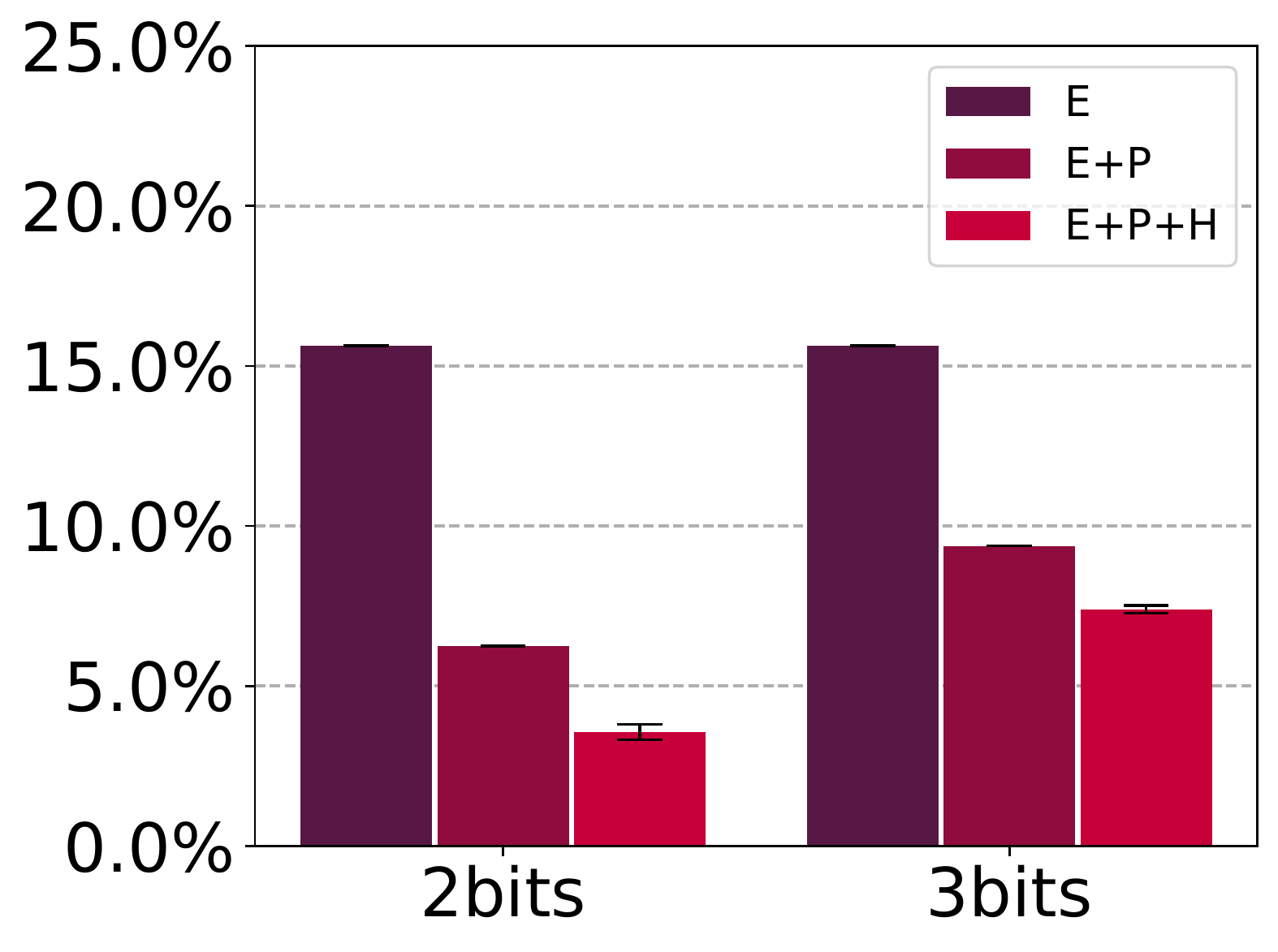}
    }
    \subfigure[MF on Jester.] { 
        \label{fig:e:size:h}     
        \includegraphics[width=0.235\linewidth]{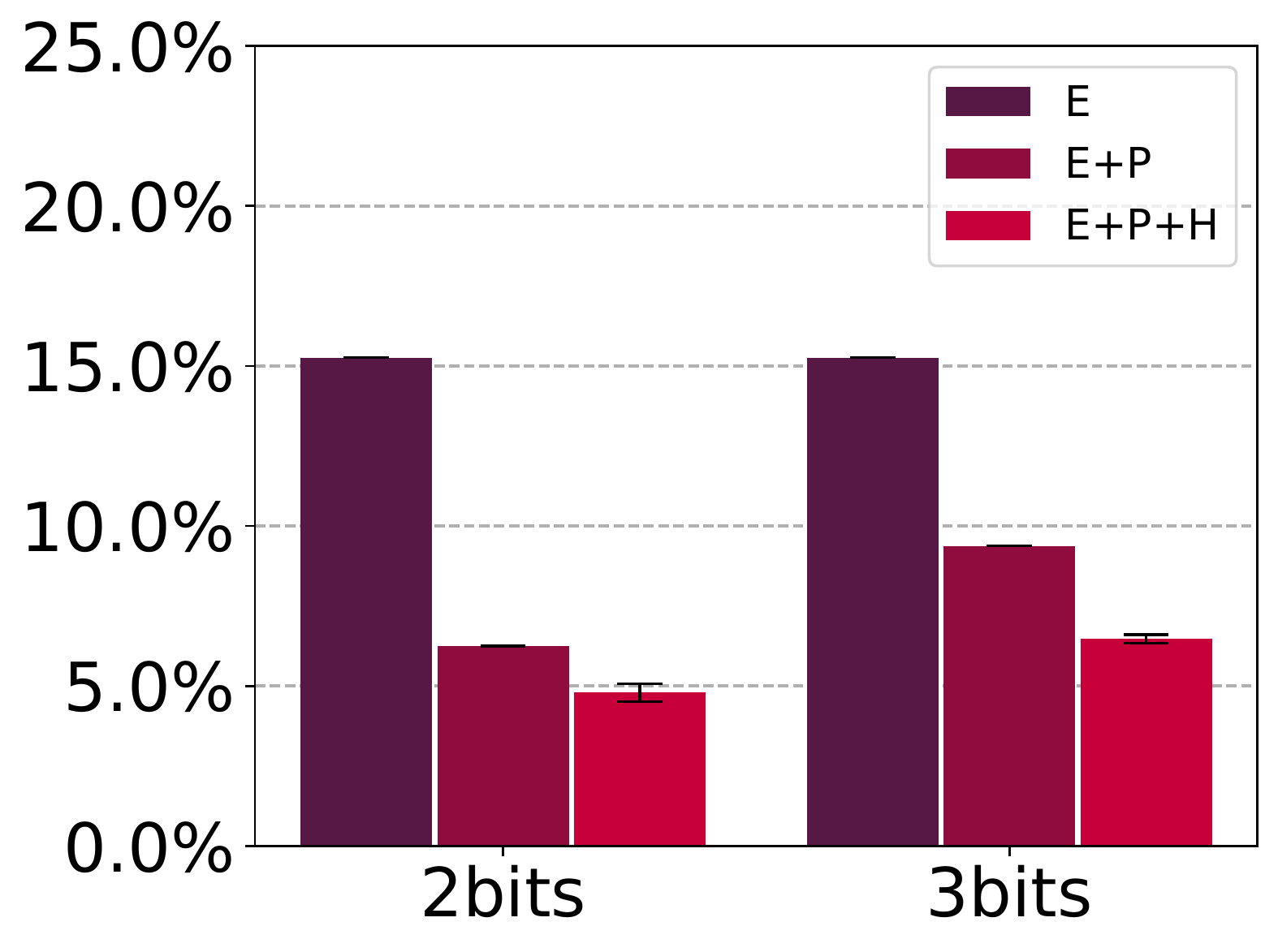}
    }
\caption{{\bf The compression ratio with different compression methods.} The x-axis denotes the bits count used in priority promotion, and the y-axis is the ratio of the checkpoint size after compression over the one before compression. E, P, H denote ``exponent-base quantization'', ``priority promotion'', and ``Huffman coding'', respectively.}     
\label{fig:e:size}     
\end{figure*}

\begin{figure}[t]
    \centering    
    \subfigure[MLR on MNIST.] {
        \label{fig:a}     
        \includegraphics[width=0.4\textwidth]{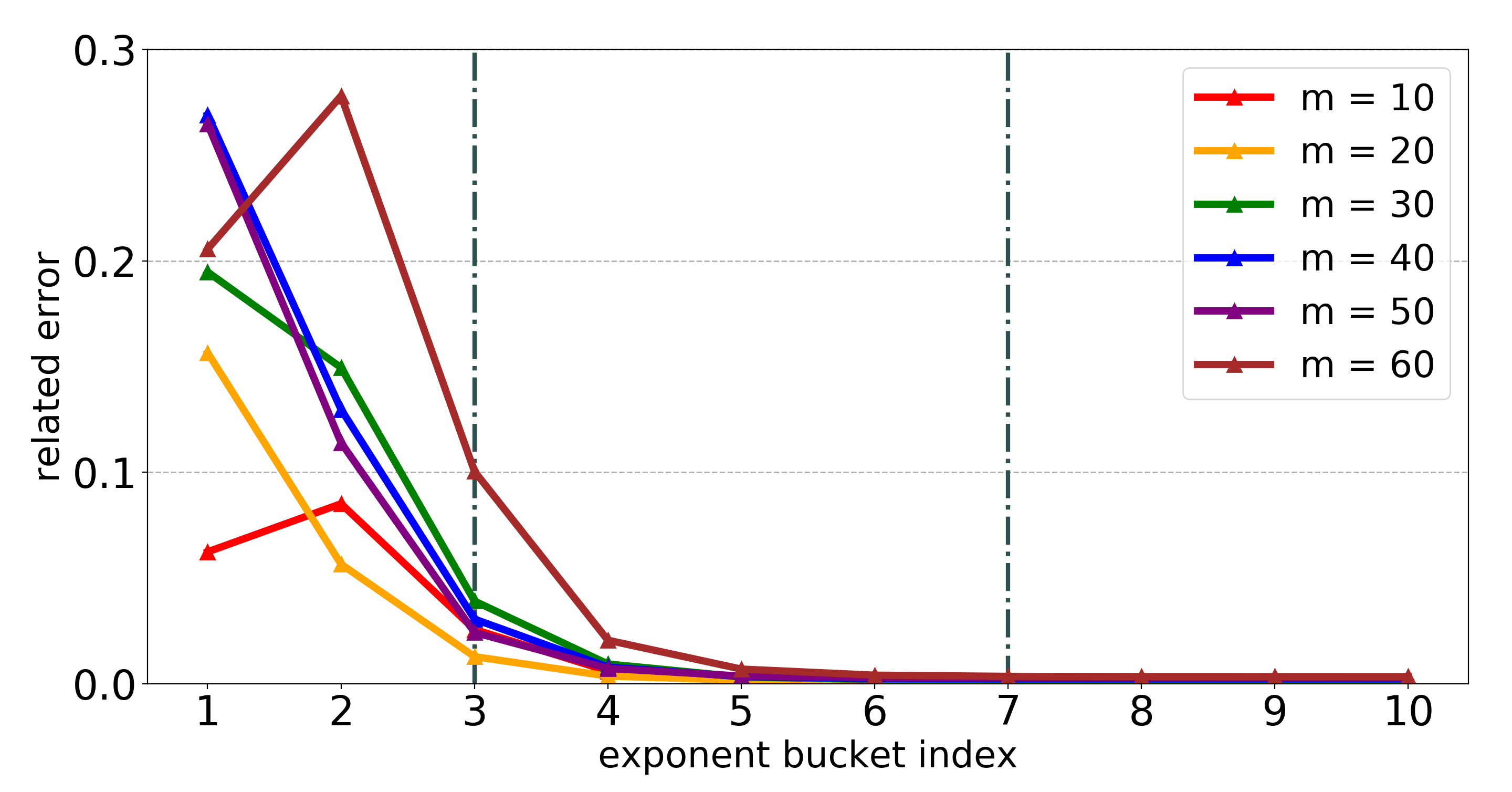}
    }     
    \subfigure[MLR on FashionMNIST.] { 
        \label{fig:b}     
        \includegraphics[width=0.4\textwidth]{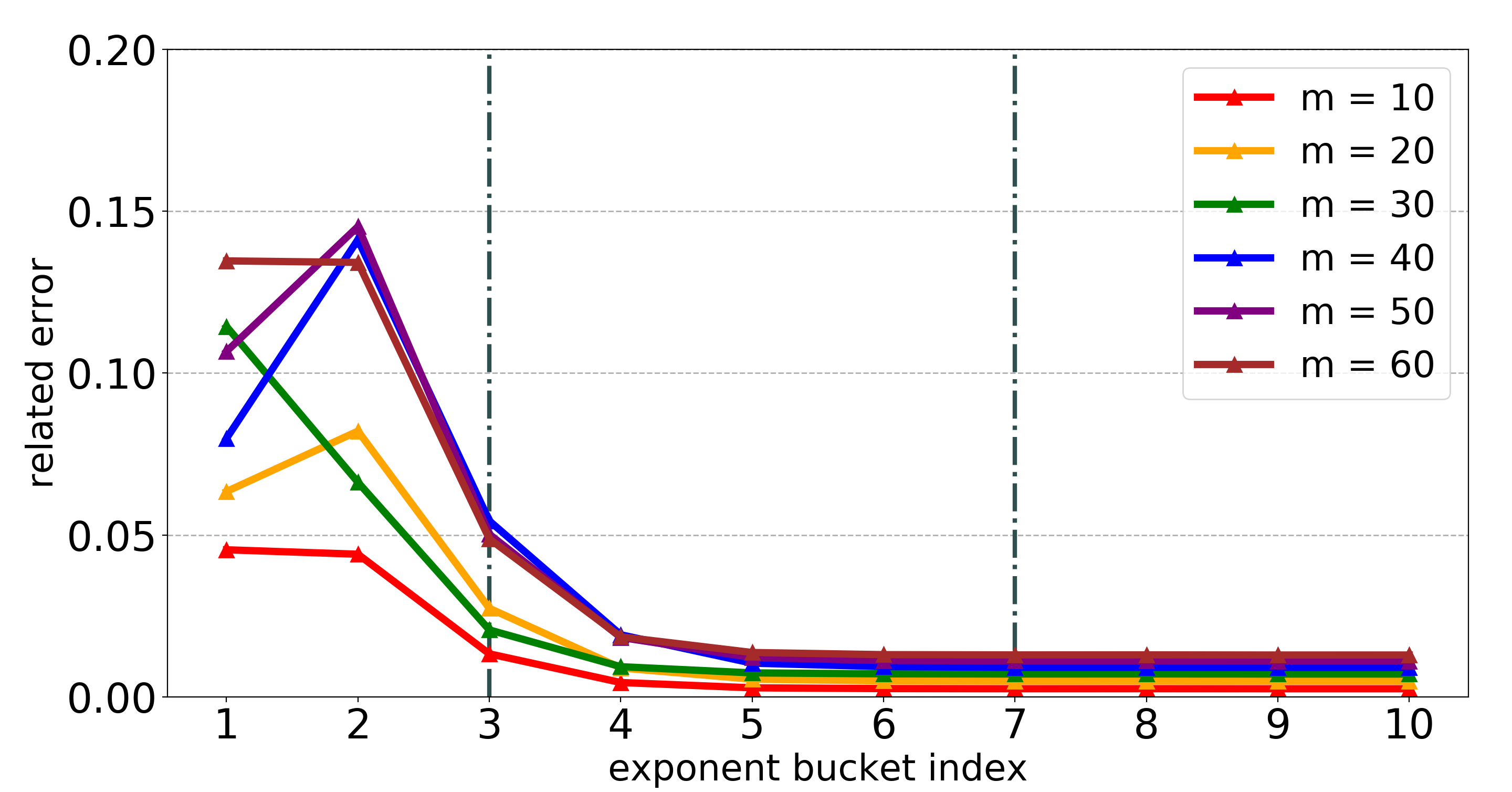}
    }
\caption{{\bf Evaluation on the priority of each exponent bucket.} The x-axis denotes the id of the exponent bucket that is deleted. The y-axis shows the relative error to the ground-truth.}
\label{fig:pp_test}     
\end{figure}

\subsection{Recovery/Rework Cost Comparison}
This section evaluates the recovery (or rework) cost of \projectname, particularly comparing it to SCAR~\cite{qiao2018fault} and a TOPN mechanism\footnote{Rework (or recovery) cost is defined as the number of iterations from $\tilde \mmu_{t}$ to $\mmu_{t}$. All methods share the same SGD computation cost for each iteration.}.

To evaluate their rework costs fairly,  we use the same checkpoint size (update size) for all three methods. Two checkpoint sizes are tested: 5\% and 10\% of the full checkpoint size\footnote{Full checkpoint stores all model parameters after a specific iteration.}. These checkpoint sizes can be set directly for SCAR and TOPN. However, \projectname's size is determined by the data distribution and thus changed dynamically. To address this issue, \projectname employs 2-bit and 3-bit priority promotion that control its checkpoint size at 5\% and 10\%. Figure~\ref{fig:e:size} reports more details of \projectname's checkpoint size information.

Figure~\ref{fig:e:overall} compares the rework cost of three methods, SCAR, TOPN, and \projectname, showing that \projectname incurs the lowest rework cost for all ML applications and datasets among them. 
For the 5\% checkpoint test case, \projectname outperforms SCAR by  $2.88\times$-$5.77\times$, and TOPN by $2.17\times$-$4.06\times$, respectively. With 10\% checkpoint size, \projectname outperforms SCAR by $1.9\times$-$4.82\times$, and outperforms TOPN by $1.52\times$-$2.17\times$, respectively. 

In addition, comparing two checkpoint sizes ($5\%$ v.s. $10\%$), \projectname results in more stable rework cost as the checkpoint size decreasing. 
For example, decreasing the checkpoint size from $10\%$ to $5\%$, \projectname has a negligible rework cost increase on {\tt LeNet} with {\tt MNIST} (Figure~\ref{fig:e:overall:b}) and {\tt AlexNet} (Figure~\ref{fig:e:overall:c}, \ref{fig:e:overall:g}). It does not have any rework cost change for other cases. In contrast, SCAR and TOPN increase $1.6\times$ rework cost on average as the checkpoint size changing from $10\%$ to $5\%$.


\subsection{\projectname Compression Effect Breakdown}
\label{section:size}

This section evaluates and analyzes the compression effect of different approaches mentioned before, exponent-base quantization ({\tt E}), priority promotion ({\tt P}), and Huffman coding ({\tt H}). Figure~\ref{fig:e:size} reports the compression ratios with 2-bit and 3-bit priority promotion. With all compression approaches, the ultimate checkpoint sizes ({\tt E+P+H}) are all below 5\% with 2-bits, and below 10\% with 3-bits over the uncompressed full checkpoint, i.e., the compression rates are above $20\times$ and $10\times$, respectively. 

Exponent-base quantization yields a compression ratio of $85\%$ on average. 
It proves that the exponent parts of all parameters in $\delta$ span across a small range of all values that can be represented by single precision floating-point. $15\%$ also indicates that the bucket number $k < 2^5$, because the average bucket number can be estimated as $k = 2^{(32 \times 15\% = 4.8)}$, where $32$ is the width of single precision floating-point. Priority promotion brings $9.26\%$ extra compression ratio on average for 2-bit and $6.23\%$ for 3-bit. For most cases, priority promotion with smaller bits yields more benefits for Huffman coding except MF (Figure~\ref{fig:e:size:d}, \ref{fig:e:size:h}). This is because MF's parameters are sparse, thus Huffman coding can reach a sufficient compression ratio without aggressive priority promotion. Across all models (and datasets), Huffman coding brings $2\%$ extra compression ratio with 2-bits priority promotion, and $1.6\%$ with 3-bits one on average.

\subsection{The Effectiveness of Priority Promotion}

This section further discusses the effectiveness of priority promotion. It aims to prove that priority promotion is able to save the majority of high priority parameters. We prove it by showing the exponent buckets result in a larger impact on the model state when their represented unique values are further from 0 (i.e., $e$ is larger). 

Assume $\delta$ is calculated from one state $\mmu_{\theta}$ to another for $m$ iterations. Then, $\delta_m^i$ is created by setting the parameters in the $i$-th exponent bucket to $0$. The ground truth is calculated as $V_{gt} = L(\mmu_{\theta}+\delta_m)$ where $L(x)$ denotes the loss function.
Then the relative error is calculated as:
\begin{equation}
    E_m^i = \frac{\left\|V_{gt}-L(\mmu_{\theta}+\delta_m^i)\right\|_2}{V_{gt}}
\end{equation}
Figure~\ref{fig:pp_test} reports the result of {\tt MLR} with $m = 10n, n \in [1, 6]$. Both datasets ({\tt MNIST} and {\tt FashionMNIST}) on varied $m$ prove that the elements in the buckets with the top-n largest distance impact more on the model (denotes as a higher relative error when the bucket represented value is set to 0).  

In addition, it is possible to preserve all {\em important} buckets with only a small number of index bits. For example, using 2-bit priority promotion (4 buckets with the last bucket storing 0) can easily preserve the most important buckets, and using 3-bit (8 buckets) can preserve all effective buckets. This result explains why priority promotion can compress the checkpoint with negligible accuracy loss.

\begin{figure}[t]
 \centering
  \includegraphics[width=0.4\textwidth]{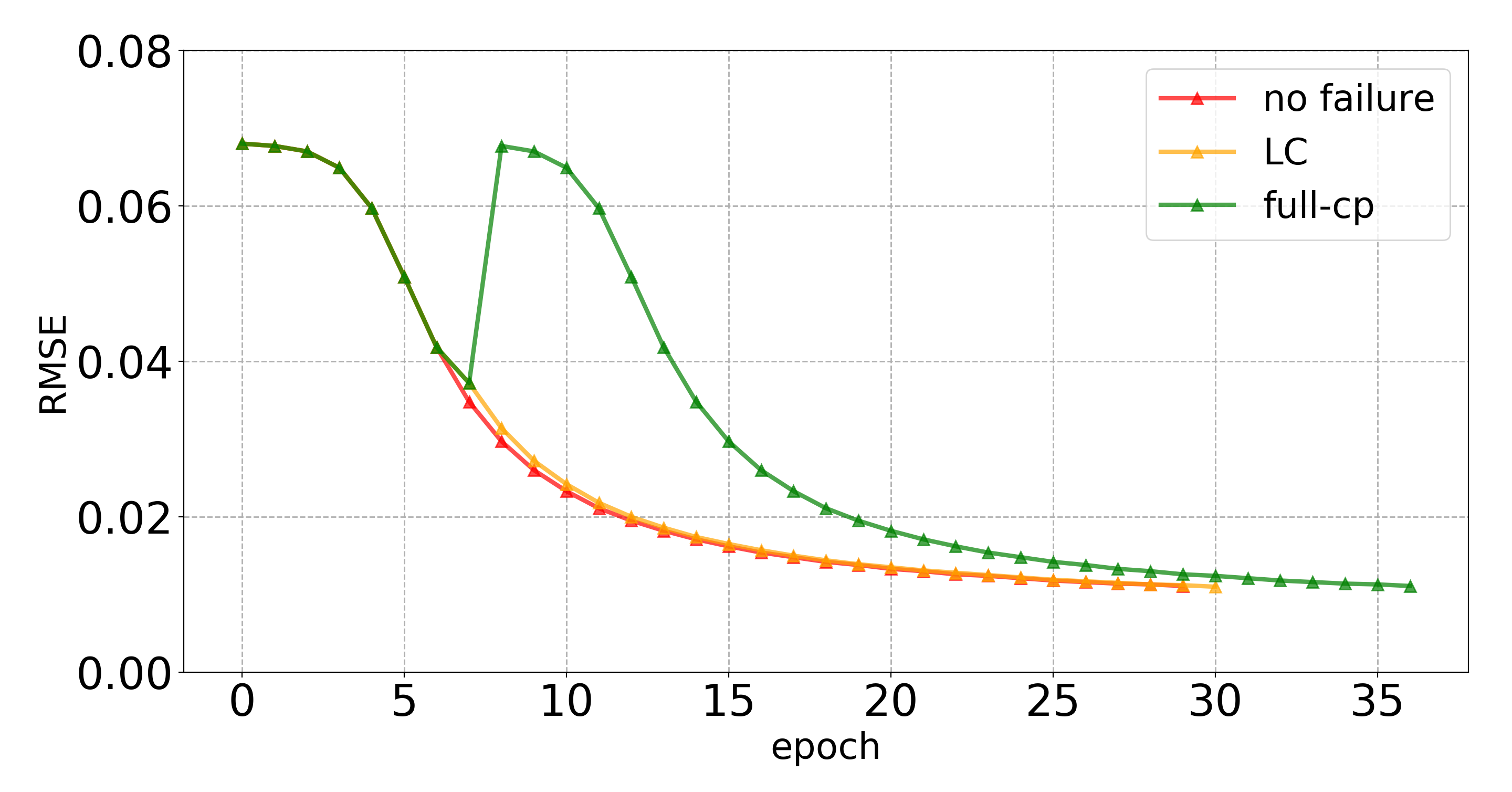}
   \caption{{\bf MF on MovieLens25M.} The x-axis denotes the iteration and the y-axis is the model's RMSE (Root Mean Square Error).}
   \label{f:bigdata}
\end{figure}

\subsection{A Case Study on \projectname's Overhead}

This section evaluates \projectname's execution overhead and overall impact on the model execution using a case study, i.e., training {\tt MF} on {\tt MovieLens25M}~\cite{harper2015movielens} dataset. Each iteration costs 91 seconds on average. \projectname employs 3-bit priority promotion, resulting in a checkpoint size below $10\%$ (of the uncompressed full checkpoint size). 
Default approach creates a full checkpoint every 10 iterations. A failure is triggered at the 7-th iteration. 

Figure~\ref{f:bigdata} reports the result.
\projectname only incurs one extra iteration than the normal execution without any failure to convergence, and saves 6 iterations compared to the full checkpoint method, i.e., saving 546 seconds execution time. \projectname introduces only less than 4 seconds (i.e., around $4\%$) overhead for each iteration, which is  negligible.

\section{Related Work}\label{sec:related}

Fault-tolerance is a key fundamental support for ML systems. Li et al.~\cite{li2014scaling} propose a runtime parameter replication approach for recovery.  Tensorflow~\cite{abadi2016tensorflow} employs periodic checkpoint to save the model state. Other efforts like~\cite{harlap2017proteus, qiao2018litz} aim to support strong consistency semantics. In contrast, our work relaxes the consistency guarantee of checkpoint based on the self-correcting behavior of ML applications. With a set of lossy compression mechanisms, our work can afford high frequent checkpoints, resulting in low rework cost and fine-grained model state recovery. Similarly, Qiao et al.~\cite{qiao2018fault} also propose a fault-tolerant solution (SCAR in our evaluation) based on weak consistency by partially updating parameters. SCAR is potential to store redundant information during checkpointing according to our evaluation, and our work aims to eliminate such redundancy by selectively saving the distance between two states. 

Model compression has been proposed to reduce model storage space and accelerate model execution time, simultaneously. Weight pruning and weight quantization are two important categories of model compression.

 Some popular weight pruning techniques closely related to our work are summarized as follows. Guo et al.~\cite{guo2016dynamic}
 present a dynamic network surgery approach with on-the-fly connection pruning to reducing the network complexity. Dai et al.~\cite{dai2019nest} combine the growth and the pruning phases in training to generate compact DNN architectures.  Han et al.~\cite{han2015learning} design Deep Compression, a model compression approach by combining pruning, quantization, and Huffman coding. Mao et al.~\cite{mao2017exploring} carefully explore the impact of varied pruning granularity on model accuracy and propose a coarse-grained weight pruning approach. All effort above aims to prune model weights without compromising accuracy. Different from them, our work eliminates the redundancy between two checkpoints and reduces the rework cost during recovery by designing a reliable coding scheme working throughout the entire dynamic process of learning.

Weight quantization is also widely used for model compression. BinaryConnect~\cite{courbariaux2015binaryconnect} introduces the binary weight for replacing multiplication by addition and subtraction.  Binarized Neural Networks~\cite{courbariaux2016binarized} also use  binary weights and activations to accelerate computation. Park et al.~\cite{park2017weighted} propose a clustering method based on weighted entropy for weight quantization. Leng et al.~\cite{leng2018extremely} formulate quantization as an optimization problem and solve it by ADMM.  Our approach also employs quantization to reduce the bits of  parameters by designing a novel exponent-based quantization technique. Moreover, our approach emphasizes filtering the parameters with a new priority promotion method.
 






\section{Conclusion and Future Work}\label{sec:conclusion}
This paper presents \projectname, the {\bf first} checkpoint scheme based on lossy compression to achieve the maximal compression rate and efficient recovery simultaneously. It employs a novel two-stage quantization method consisting of exponent-based quantization and priority promotion to identify and store the most critical information for SGD to recover, and leverages Huffman coding to further benefit from the non-uniform distribution of gradient scales. Our evaluation demonstrates that \projectname achieves a compression rate up to $28\times$ and recovery speedup up to $5.77\times$ over the state-of-the-art algorithm ({\tt SCAR}). 

In the future, we plan to generalize \projectname by relaxing the assumption of SGD and equipping it with the capability of selecting checkpoint compression rates dynamically according to model and data changes.  

\section*{Acknowledgements}
We thank the anonymous reviewers for their valuable comments. This work is supported in part by NSF grants CRII-1755646, CRII-1755769, OAC-1835821, CNS-1813487 and CCF-1918757, a Google Faculty Research Award, and an AWS Machine Learning Research Award.






\bibliographystyle{icml2020}
\bibliography{icml2020}

\end{document}